%% file: main.tex
\documentclass[letterpaper]{article}
\usepackage{aaai}
\usepackage{times}
\usepackage{helvet}
\usepackage{courier}
\usepackage{amsmath,amssymb,amsthm,breqn}
\usepackage{graphicx}

\usepackage[caption=false]{subfig}
\usepackage{booktabs} 
\usepackage{tabularx}

\usepackage{amsmath,amssymb,amsfonts}
\usepackage{bm}
\usepackage{bbm}
\usepackage{hyperref}

\usepackage{cleveref}
\usepackage{xcolor}

\usepackage{amsthm}

\newtheorem{definition}{Definition}[section]

\hypersetup{pdfauthor={Masahiro Nomura, Shuhei Watanabe, Youhei Akimoto, Yoshihiko Ozaki, Masaki Onishi},%
            pdftitle={Warm Starting CMA-ES for Hyperparameter Optimization}
}
\begin{document}
%

\title{Warm Starting CMA-ES for Hyperparameter Optimization}

\author{Masahiro Nomura\thanks{These authors contributed equally in this work.}\textsuperscript{1,2}, Shuhei Watanabe\footnotemark[1]\thanks{The work was done at Artificial Intelligence Research Center, AIST.}\textsuperscript{3}, Youhei Akimoto\textsuperscript{4,5}, Yoshihiko Ozaki\textsuperscript{2,6}, Masaki Onishi\textsuperscript{2}\\
	\textsuperscript{1}CyberAgent, Inc. nomura\_masahiro@cyberagent.co.jp\\
	\textsuperscript{2}Artificial Intelligence Research Center, AIST. \\
	\textsuperscript{3}University of Freiburg.\\
	\textsuperscript{4}University of Tsukuba.\\
	\textsuperscript{5}RIKEN Center for Advanced Intelligence Project.\\
	\textsuperscript{6}GREE, Inc.
}

\maketitle
\begin{abstract}
Hyperparameter optimization (HPO), formulated as black-box optimization (BBO), is recognized as essential for automation and high performance of machine learning approaches.
The CMA-ES is a promising BBO approach with a high degree of parallelism, and has been applied to HPO tasks, often under parallel implementation, and shown superior performance to other approaches including Bayesian optimization (BO).
However, if the budget of hyperparameter evaluations is severely limited, which is often the case for end users who do not deserve parallel computing, the CMA-ES exhausts the budget without improving the performance due to its long adaptation phase, resulting in being outperformed by BO approaches.
To address this issue, we propose to transfer prior knowledge on similar HPO tasks through the initialization of the CMA-ES, leading to significantly shortening the adaptation time.
The knowledge transfer is designed based on the novel definition of task similarity, with which the correlation of the performance of the proposed approach is confirmed on synthetic problems.
The proposed warm starting CMA-ES, called WS-CMA-ES, is applied to different HPO tasks where some prior knowledge is available, showing its superior performance over the original CMA-ES as well as BO approaches with or without using the prior knowledge.
\end{abstract}

\input{manuscripts/1_Introduction}
\input{manuscripts/2_Background}
\input{manuscripts/3_Warm_Starting}
\input{manuscripts/4_Experiments_Synthetic}

\input{manuscripts/5_Experiments_HPO}

\input{manuscripts/6_Related_Work_Discussion}
\input{manuscripts/7_Conclusion}


\section*{Acknowledgements}
The authors thank Shota Yasui, Yuki Tanigaki, Yoshiaki Bando for valuable feedback and suggestion.
This paper is based on the results obtained from a project commissioned by the New Energy and Industrial Technology Development Organization (NEDO). Computational resource of AI Bridging Cloud Infrastructure (ABCI) provided by National Institute of Advanced Industrial Science and Technology (AIST) was used.
\bibliography{ref}
\bibliographystyle{aaai}

\clearpage
\input{manuscripts/8_Appendix}

\clearpage

\end{document}

%% file: manuscripts/1_Introduction.tex
\section{Introduction}
\label{sec:intro}
Hyperparameter optimization (HPO) is an essential for achieving effective performance in a wide range of machine learning algorithms~\cite{feurer2019hyperparameter}.
HPO is formulated as a black-box optimization (BBO) problem because the objective function of the task of interest (referred to as the target task) cannot be described using an algebraic representation in general.
One way to accelerate the optimization for HPO on the target task is to exploit results from a related task (referred to as the source task).
This \emph{transfer learning} setting on HPO often appears in practical situations and is actively studied in HPO literature~\cite{vanschoren2019meta}.

The covariance matrix adaptation evolution strategy (CMA-ES)~\cite{hansen2001completely,hansen2016cma} is one of the most powerful methods for BBO and has a high degree of parallelism.
The CMA-ES facilitates optimization by updating the parameters of the multivariate Gaussian distribution (MGD).
Subsequently, the CMA-ES samples candidate solutions, which can be evaluated in parallel, from the MGD.
It has been applied widely in practice, including in HPO often under parallel evaluation settings~\cite{loshchilov2016cma,friedrichs2005evolutionary,watanabe2014black}.
The CMA-ES is particularly useful for solving \emph{difficult} BBO problems such as non-separable, ill-conditioned, and rugged problems~\cite{rios2013derivative};
furthermore, it has shown the best performance among more than $100$ optimization methods for various BBO problems~\cite{loshchilov2013bi} with moderate to large evaluation budgets ($> 100 \times$ the number of variables).

However, the CMA-ES does not necessarily outperform Bayesian optimization (BO)~\cite{frazier2018tutorial} in the context of HPO, in which the evaluation budget is severely limited~\cite{loshchilov2016cma}.
This is because the CMA-ES requires a relatively long adaptation phase to sample solutions into promising regions, especially at the beginning of optimization.
Thus, the CMA-ES has received much less attention in the context of HPO, despite the excellent performance verified empirically in BBO.

In this work, to address the inefficiency of the CMA-ES when the evaluation budget is severely limited, we introduce a simple and effective warm starting method WS-CMA-ES.
This warm starting strategy can shorten the CMA-ES adaptation phase significantly by utilizing the relationship between source and target tasks.

We first define a promising distribution in the search space and task similarity.
The proposed method is designed to perform successful warm starting when the defined task similarity between a source task and a target task is high.
To warm-start the optimization, we estimate a promising distribution of the source task.
The mean vector and the covariance matrix that are the parameters of the MGD in the CMA-ES are initialized by minimizing the Kullback--Leibler (KL) divergence between the MGD and the promising distribution.
In this study, we performed experiments on synthetic and HPO problems for several warm starting settings.
In particular, we have compared the proposed method with the original CMA-ES, Gaussian process expected improvement (GP-EI)~\cite{NIPS2012_4522}, tree-structured Parzen estimator (TPE)~\cite{bergstra2011algorithms}, multi-task Bayesian optimization (MTBO)~\cite{swersky2013mtbo}, and multi-task BOHAMIANN (MT-BOHAMIANN)~\cite{springenberg2016bayesian}.


In summary, the contributions of this work are as follows:
\begin{itemize} 
    \item We formally defined a promising distribution and task similarity to give us the insight required to design a warm starting strategy for the CMA-ES.
    \item We proposed a warm starting method called the WS-CMA-ES that speeds up HPO by reducing the adaptation phase of the CMA-ES.
    \item We demonstrated that the performance of WS-CMA-ES is correlated with the defined task similarity through numerical experiments.
    \item We verified by synthetic problems that the WS-CMA-ES is more effective than naive warm starting methods even when a source task and a target task are not very similar.
    \item We demonstrated that the WS-CMA-ES converges faster than the existing methods for HPO problems.
\end{itemize}

%% file: manuscripts/2_Background.tex
\section{Background}
In this study, we considered the following BBO problem: minimizing a black-box function $f : \mathcal{X} \to \mathbb{R}$ over a compact measurable subset $\mathcal{X} \subseteq \mathbb{R}^d$, where $d$ is the number of variables.
Let $\Lambda_{\rm Leb}$ be the Lebesgue measure on $\mathcal{X}$.
In HPO, a solution $x \in \mathcal{X}$ corresponds to one hyperparameter setting, and $f(x)$ is generally a validation error of the trained model.

\subsection{CMA-ES}
\label{sec:cma}

The CMA-ES is a BBO method that uses an MGD $\mathcal{N}(m, \Sigma)$, wherein $m \in \mathbb{R}^d$, $\Sigma \in \mathbb{R}^{d \times d}$ is a positive-definite symmetric matrix.
This algorithm generates $\lambda$ solutions following the MGD and evaluates each solution in every iteration, which is defined as a generation.
The mean vector $m$ and the covariance matrix $\Sigma$ are updated according to the ranking of the solutions in the latest generation, and the CMA-ES learns to sample solutions from the promising region.
\footnote{Note that we followed the standard formulation of the CMA-ES.
Therefore, $\Sigma$ was decomposed into $\Sigma = \sigma^2 C$ where $\sigma > 0$ and $C \in \mathbb{R}^{d \times d}$.}
The update of the CMA-ES is strongly related to the natural gradient descent~\cite{akimoto2010bidirectional,ollivier2017information};
$m$ and $\Sigma$ in the CMA-ES are updated to decrease the expected evaluation value.
For more details, see the CMA-ES tutorial~\cite{hansen2016cma}.%
\footnote{Among several versions of the CMA-ES, we use the recent standard version described in~\cite{hansen2016cma}.}

The CMA-ES is invariant with order-preserving transformations of the objective function because the CMA-ES uses only a ranking of solutions, not the evaluation value itself.
In addition, the CMA-ES has the affine invariance to the search space.
These invariances allow us to generalize the successful empirical results of the CMA-ES to a more wide range of problems~\cite{hansen2014principled}.

\subsection{CMA-ES for Hyperparameter Optimization}
\label{sec:cma_hpo}
The invariances mentioned above make the CMA-ES suitable for HPO.
For example, when transferring HPO to different \emph{dataset} or different \emph{objectives}, the scale of each objective may significantly vary.
However, the CMA-ES is robust to such a heterogeneous scale owing to the use of rank, not the evaluation value itself.
Further, hyperparameters are often dependent on each other, such as the batch size and the learning rate in deep neural networks~\cite{keskar2016large,smith2017don}.
The CMA-ES can address this dependency by learning the covariance matrix appropriately.
Indeed, Loshchilov and Hutter \cite{loshchilov2016cma} reported that the CMA-ES outperformed BO in HPO when a moderate evaluation budget was available.
However, the evaluation budget is often severely limited and is insufficient for the CMA-ES to adapt the covariance matrix, particularly for the end users whose computational resources are limited.
In such cases, the CMA-ES does not yield better solutions than other approaches such as BO approaches~\cite{loshchilov2016cma}.

The possible reason for the lower performance of the CMA-ES is a long adaptation phase of the covariance matrix;
we explain the reason below.
The CMA-ES attempts to adapt the covariance matrix to approximate the shape of the level set of the objective function by that of MGD.
In the case of a convex quadratic objective function, the covariance matrix approximates the inverse Hessian of the function.
Once the covariance matrix is well-adapted, the CMA-ES exhibits a linear convergence, where the convergence speed is as high as the one for the spherical function, $f(x) = \|x\|^2$.
However, as the degree of freedom of the covariance matrix is $\Theta(d^2)$, the learning rate for the covariance matrix update is set to $\Theta(1/d^2)$ by default for stability.
Therefore, $\mathcal{O} (d^2)$ iterations are required to adapt the covariance matrix.

Two approaches can be considered to mitigate this problem.
One is increasing $\lambda$, which is the number of solutions per iteration, and evaluating them in parallel.
The number of iterations for the adaptation decreases as $\lambda$ increases~\cite{akimoto2019diagonal}.
However, this approach is useful only for those users who can afford parallel computational environments.
The other approach is employing variants of the CMA-ES with a restricted covariance matrix model, such as the diagonal model~\cite{ros2008simple}.
Because the covariance matrix model has few degrees of freedom, the learning rate can be set higher, thereby accelerating the adaptation, while compromising rotational invariance.
Hence, we also propose another method which uses a restricted matrix model, in addition to the proposed method with the full covariance matrix.

%% file: manuscripts/3_Warm_Starting.tex
\section{Warm Starting CMA-ES}
We consider a transfer HPO setting, where we have pairs ({\it hyperparameter, performance}) on a source task.
This setting often appears in the practical use of HPO~\cite{vanschoren2019meta}.
The original CMA-ES and other variants aiming to mitigate the problem of the long adaptation phase, which are described in Section \ref{sec:cma_hpo}, do not have any mechanism to exploit such observational data on the source task.

In this work, we propose a simple and effective warm starting CMA-ES (WS-CMA-ES).
First, we construct the definitions of a promising distribution and a task similarity.
Next, the details of WS-CMA-ES are given.

\subsection{Definition of Task Similarity}
\label{sec:similarity}
To define task similarity, it is necessary to identify which parts of the objective function characterize the task.
Because the goal of optimization is to identify the best solution, one possible definition of a task feature is a promising distribution, which represents the regions wherein promising solutions exist with a higher probability.
Herein, we define the {\it $\gamma$-promising distribution} as follows:
\vspace{+2mm}
\begin{definition}[$\gamma$-Promising Distribution]
    \label{def:gamma_promising_distribution}
    Suppose that $f : \mathcal{X} \to \mathbb{R}$ is a measurable function defined over the compact measurable subset $\mathcal{X} \subseteq \mathbb{R}^d$.
    For $\gamma \in (0, 1]$, let $\mathcal{F}^{\gamma} = \{ x \in \mathcal{X} \mid f(x) \leq f^{\gamma} \}$, where $f^{\gamma}$ is defined such that $\gamma = \Lambda_{\rm Leb} (\mathcal{F}^{\gamma}) / \Lambda_{\rm Leb} (\mathcal{X})$.
    We define $\gamma$-promising distribution $P^{\gamma}$, whose probability density function $p^{\gamma}$ is defined as
    \small
    \begin{align}
    \label{eq:promising_form}
      p^{\gamma}(x) &= \frac{1}{Z} \int_{x' \in \mathcal{F}^{\gamma}} 
      \exp \Bigl( - \frac{\| x - x' \|^2}{2\alpha^2} \Bigr) dx' \enspace,
    \end{align}
    \normalsize
    where $Z = \int_{x \in \mathcal{X}} \int_{x' \in \mathcal{F}^{\gamma}} \exp \Bigl( - \frac{\| x - x' \|^2}{2\alpha^2} \Bigr) dx' dx$ and $\alpha \in \mathbb{R}^{>0}$ is a prior parameter.
\end{definition}

Our definition of the $\gamma$-promising distribution is based on two HPO problem assumptions: (1) the continuity between hyperparameters and objective function and (2) the local convexity of a promising region.

The first assumption is the continuity of the objective function.
When a hyperparameter varies slightly, its performance also changes to a small extent.
More precisely, the continuity of the objective function leads to the smoothness of the promising distribution.
Another possible definition for the promising distribution is a uniform distribution $\mathbbm{1} \{ x \in \mathcal{F}^{\gamma} \} / \Lambda_{\rm Leb} (\mathcal{F}^{\gamma})$, where $\mathbbm{1}$ is the indicator function.
An advantage of the uniform distribution is its simplicity.
However, the measure of the regions not within $\mathcal{F}^{\gamma}$ becomes $0$ when the promising distribution is based on the uniform distribution.
In other words, this distribution considers the regions in $\mathcal{F}^{\gamma}$ as promising to the same extent and considers the other regions totally unpromising.
The main problem with the uniform distribution is that it ignores the importance of the proximity of the boundaries around $\mathcal{F}^{\gamma}$.
In fact, the magnitudes of importance for the boundary regions should not fluctuate greatly depending on whether the regions are inside or beyond $\mathcal{F}^{\gamma}$.
The promising distribution should measure such slight variations of importance over the entire search space.
This condition requires the promising distribution to be smooth.
Therefore, the uniform distribution, which does not satisfy the smoothness, is not suitable for defining the promising distribution.

The second assumption is related to the local convexity of the promising region.
From the first assumption, the inside of the level set is likely to be more promising, and it naturally leads to the local convexity.
The $\gamma$-promising distribution can represent these assumptions more appropriately than the uniform distribution $\mathbbm{1} \{ x \in \mathcal{F}^{\gamma} \} / \Lambda_{\rm Leb} (\mathcal{F}^{\gamma})$.

Next, {\it $\gamma$-similarity}, which measures task similarity, is formulated using the $\gamma$-promising distribution as follows:
\vspace{+2mm}
\begin{definition}[$\gamma$-Similarity]
    \label{def:gamma_similarity}
    Suppose that $f_1,\ f_2 : \mathcal{X} \to \mathbb{R}$ are  measurable functions defined over the compact measurable subset $\mathcal{X} \subseteq \mathbb{R}^d$. Let $\gamma_1, \gamma_2 \in (0, 1]$ and let $P_i^{\gamma_i}$ be $\gamma_i$-promising distribution of $f_i$ for $i = 1, 2$ defined in \Cref{def:gamma_promising_distribution}.
    We define $\gamma$-similarity from $f_1$ to $f_2$ as
    \begin{align}
      s(\gamma_1, \gamma_2) := D_{\rm KL}(P_{*} || P_2^{\gamma_2}) - D_{\rm KL}(P_1^{\gamma_1} || P_2^{\gamma_2}),
    \end{align}
    where $D_{\rm KL}(P || Q)$ is the KL divergence between $P$ and $Q$, and $P_{*}$ is a non-informative prior distribution.
\end{definition}

A non-informative prior distribution is used when no information is available on the objective function.
In the CMA-ES, for the search space $\mathcal{X} = [0, 1]^d$, $\mathcal{N}(0.5, 0.2^2)$ is given as an initial distribution for each variable~\cite{loshchilov2016cma}.
BO uses a uniform distribution as a non-informative prior distribution in many cases.

Intuitively, if $s(\gamma_1, \gamma_2) > 0$, i.e., $D_{\rm KL}(P_1^{\gamma_1} || P_2^{\gamma_2}) < D_{\rm KL}(P_{*} || P_2^{\gamma_2})$, the promising distribution $P_1^{\gamma_1}$ for task $1$ is closer to the promising distribution $P_2^{\gamma_2}$ for task $2$ than a non-informative prior distribution $P_*$.

\subsection{Algorithm Construction}
\label{sec:alg_ws_cma}

We assume a source task (task 1) is similar to a target task (task 2).
Hence, the $\gamma$-similarity holds $s(\gamma_1, \gamma_2) > 0$, i.e., $D_{\rm KL}(P_1^{\gamma_1} || P_2^{\gamma_2}) < D_{\rm KL}(P_{*} || P_2^{\gamma_2})$.
Note that the non-informative prior distribution $P_{*}$ is used as an initial distribution for the CMA-ES if there is no information on the source task.
Because we assume knowledge transfer from the source task, we obtain $D_{\rm KL}(P_1^{\gamma_1} || P_2^{\gamma_2}) < D_{\rm KL}(P_{*} || P_2^{\gamma_2})$.
Therefore, the CMA-ES can begin optimization from the location close to the promising region by exploiting the information for the promising region of the source task from $P_1^{\gamma_1}$ instead of $P_{*}$.
In this study, the initial parameters of the MGD were estimated by minimizing the KL divergence between the MGD and the empirical version of $P_1^{\gamma_1}$.
The empirical version of $P_1^{\gamma_1}$ uses Gaussian mixture models (GMM) as shown in Eq.~(\ref{eq:gmm}).

This method transfers prior knowledge as follows.
First, the top $\gamma \times 100$\% solutions are selected from a set of solutions on a source task. 
Second, a GMM, i.e., a promising distribution of the source task, is built using the solutions selected above.
Finally, the parameters of the MGD are initialized via the approximation of the GMM.
Further details of each operation are described in the next section.

\subsection{Details of Each Operation}
\subsubsection{Estimation of a Promising Distribution of a Source Task}
\label{sec:estimate_promising_regions}
Let $N$ be the number of observations in a source task.
Based on the definition of $\gamma$-promising distribution, a probability density function $p(x)$ that represents a promising distribution of the source task is estimated using the top $\gamma \times 100$ \% solutions as follows:
\begin{align}
    \vspace{-2mm}
    \label{eq:gmm}
    p(x) = \frac{1}{N_{\gamma}} \sum_{i=1}^{N_{\gamma}} \mathcal{N}(x \mid x_{i}, \alpha^2 I),
\end{align}
where $\alpha \in \mathbb{R}^{>0}$ is a prior parameter, $I \in \mathbb{R}^{d \times d}$ is the identity matrix, $N_{\gamma} = \lfloor \gamma \cdot N \rfloor$, and $x_{i}$, which is an observation in the source task, is sorted so that $f(x_{1}) \leq f(x_{2}) \leq \cdots \leq f(x_{N_{\gamma}}) \leq \cdots \leq f(x_{N})$.
The robustness of these two parameters, $\gamma$ and $\alpha$, is shown in Appendix \ref{supple_sec:hp_ws_cma}.

\subsubsection{Transferring Prior Knowledge to the CMA-ES}
\label{sec:initial_param_cma}
Based on the aformentioned promising distribution definition, we introduced the estimation method for the initial parameters of the MGD for the CMA-ES.
The initial parameters were determined by minimizing the KL divergence between the promising distribution $p(x)$ and the MGD $q(x) = \mathcal{N}(x \mid m, \Sigma)$.

Based on Theorem 3.2 and Eqs.~(2)--(4) of~\cite{runnalls2007kullback}, we can easily identify the parameters that minimize the KL divergence as follows:
\begin{align}
    m^{*} &= \frac{1}{N_{\gamma}} \sum_{i=1}^{N_{\gamma}} x_{i}, \\
    \label{eq:sigma_mae}
    \Sigma^{*} &= \alpha^2 I + \frac{1}{N_{\gamma}} \sum_{i=1}^{N_{\gamma}} (x_{i} - m^{*}) {(x_{i} - m^{*})}^{\top}.
\end{align}

We can observe that Eq.~(\ref{eq:sigma_mae}) agrees with the formula for the maximum a posteriori estimation~\cite{bishop2006pattern}, which implies that the first term in Eq.~(\ref{eq:sigma_mae}) has the effect of the regularization.
We can also derive a variant that restricts the covariance matrix to a diagonal: $\Sigma^{*} = {\rm diag}(l_1, \cdots, l_d)$.
For $j \in \{ 1, \cdots, d \}$, it can be easily calculated as follows, considering the independence of the variables: 
\begin{align}
l_j = \alpha^2 + \frac{1}{N_{\gamma}} \sum_{i=1}^{N_{\gamma}} ([x_{i}]_j - [m^{*}]_j)^2,
\end{align}
where $[x]_j$ denotes the $j$-th element of the vector $x$.
We call this restricted variant WS-sep-CMA-ES; its performance is validated in Section \ref{sec:naive_transfer}.

%% file: manuscripts/4_Experiments_Synthetic.tex
\section{Experiments on Synthetic Problems}
\label{sec:exp_gamma_similarity}

\subsection{Performance Depending on Task Similarity}
As defined in the previous section, WS-CMA-ES is expected to achieve faster convergence on problems with higher $\gamma$-similarity (i.e. $s(\gamma_1, \gamma_2) > 0$).
To confirm this correlation, we measured the $\gamma$-similarity and the performance of WS-CMA-ES using two synthetic problems:
\begin{itemize}
    \vspace{0mm}
    \item {\bf Sphere Function}: $f(x) = (x_1 - b)^2 + (x_2 - b)^2$
    \vspace{0mm}
    \item {\bf Rotated Ellipsoid Function}: $f(x) = f_{\rm ell} (R x)$
    \vspace{0mm}
\end{itemize}
where $f_{\rm ell}(x) = (x_1 - b)^2 + 5^2 (x_2 - b)^2$, $R \in \mathbb{R}^{2 \times 2}$ is a rotation matrix rotating $\pi / 6$ around the origin, and $b$ is the coefficient for each problem.
The sphere function is a simple problem.
On the other hand, the characteristics of the rotated ellipsoid function are non-separable and ill-conditioned.
Non-separable characteristic is related to the dependencies between the variables, and the ill-conditioned characteristic is that the contribution to the objective function varies widely depending on each variable.

We optimized each synthetic problem using two methods: the WS-CMA-ES and the original CMA-ES.
The target task for each problem is the function with a coefficient $b=0.6$.
As prior knowledge for both settings, we evaluated each function with a coefficient from $b=0.4,\cdots, 0.8$ in increments of $0.1$.
In other words, we optimized the synthetic problems by the WS-CMA-ES using each prior knowledge for each case (for $b=0.4,0.5,\cdots,0.8$).
Each optimization was performed $20$ times.
We employed $\mathcal{N}(0.5, 0.2^2)$, which is a non-informative distribution used in Definition \ref{def:gamma_similarity}, as an initial distribution for each variable in the CMA-ES.
In all the experiments, $\alpha$ and $\gamma$ in WS-CMA-ES were set to $0.1$ for each.
For more details, see Appendices A and B.

\begin{figure}[t]
\vspace{-6mm}
    \centering
    \hspace*{\fill}
    \subfloat[][Sphere Function \label{subfig:best_mean_diff}]{
        \includegraphics[width=0.47\linewidth]{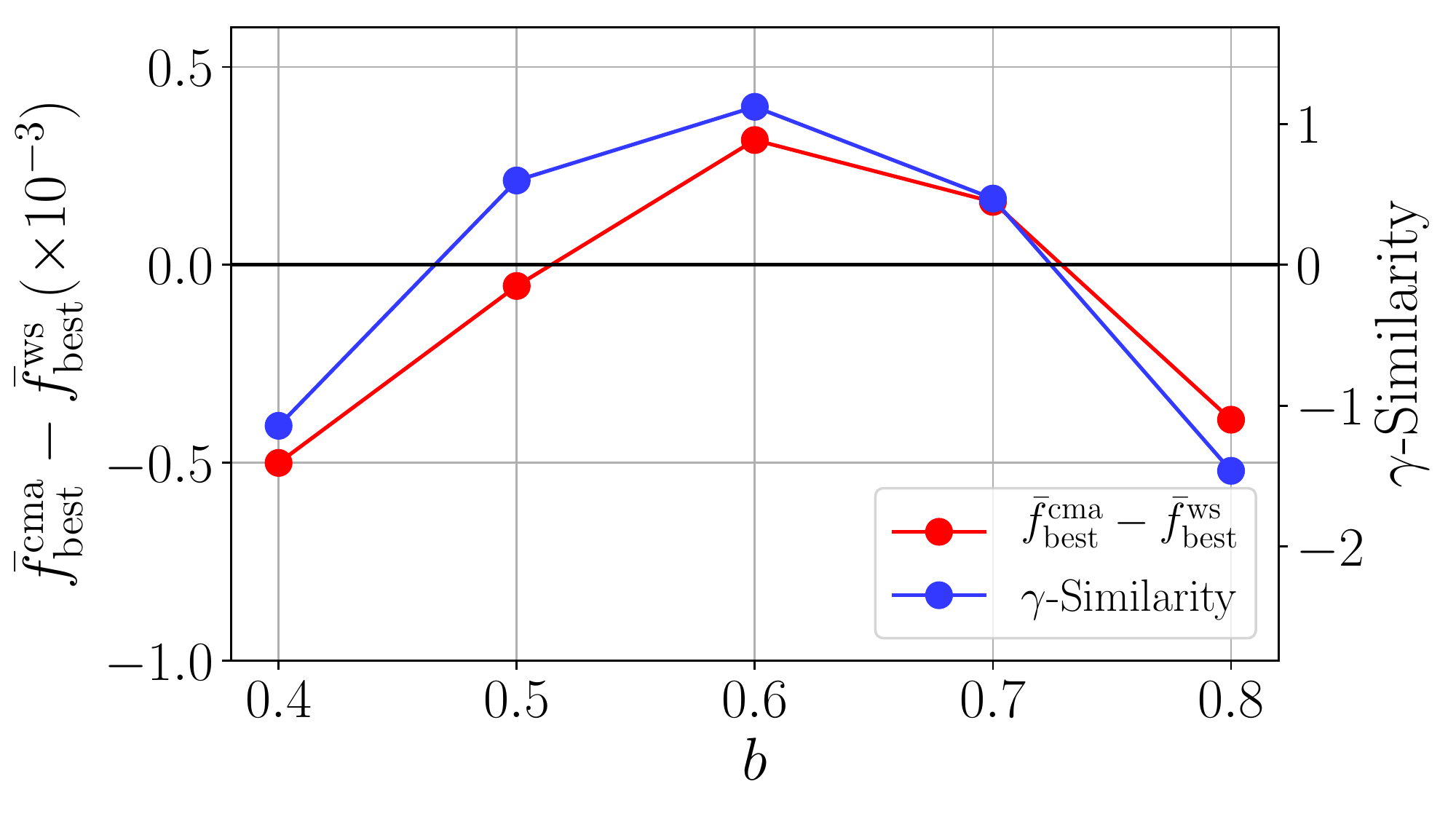}
    }
    \hspace*{\fill}
    \subfloat[][Rotated Ellipsoid Function \label{subfig:exp_gamma_similarity}]{
        \includegraphics[width=0.47\linewidth]{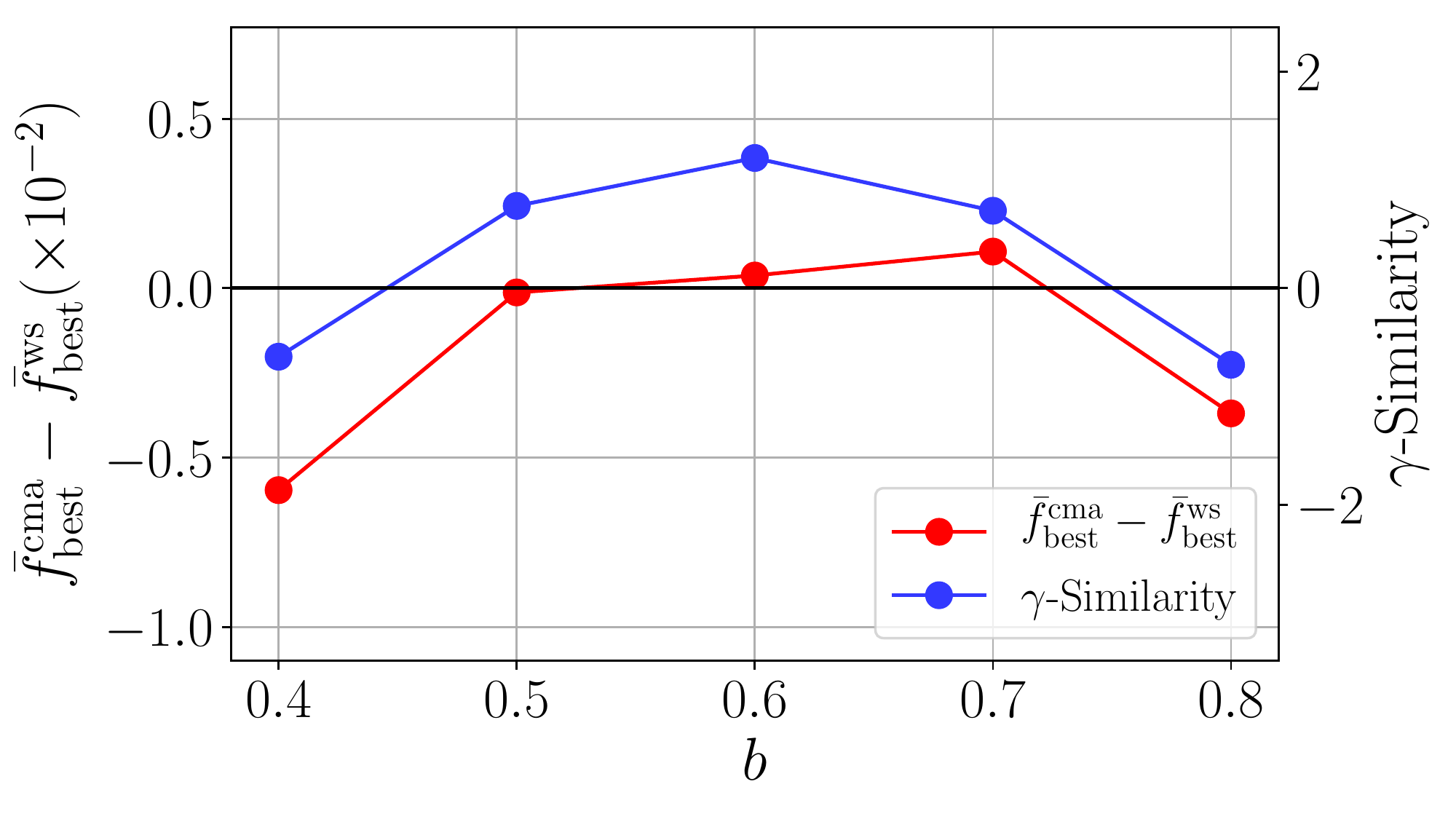}
    }
    \hspace*{\fill}
    \caption{Results of the experiments to confirm the correlation between $\gamma$-similarity and performance. 
    The horizontal axis represents the prior parameter $b$ used for each warm starting setting where the prior knowledge was the result with $b=0.6$.
    The vertical axis for red and blue lines denote the subtraction $\bar{f}_{\rm best}^{\rm cma} - \bar{f}_{\rm best}^{\rm ws}$ of the mean of the best evaluation value in the CMA-ES and the WS-CMA-ES ($20$ runs for each) and $\gamma$-similarity in \Cref{def:gamma_similarity}, respectively. $\bar{f}_{\rm best}^{\rm cma} - \bar{f}_{\rm best}^{\rm ws} > 0$ implies that the result of the WS-CMA-ES is better than that of the CMA-ES.}
    \label{fig:exp_performance_vs_similarity}
\end{figure}

The results are presented in Figure~\ref{fig:exp_performance_vs_similarity}.
To visualize the correlation between the performance improvement and the $\gamma$-similarity, we measured the $\gamma$-similarity between the cases of $b=0.4,\cdots, 0.8$ and $b=0.6$ and plotted it along with the results.
In both settings, the variation of the $\gamma$-similarity almost corresponds to that of the improvement achieved by WS-CMA-ES compared with the CMA-ES with respect to the value $b$.
In brief, WS-CMA-ES successfully transferred prior knowledge when a source task resembled the target task in terms of $\gamma$-similarity;
in contrast, when the task similarity was low, the WS-CMA-ES did not perform well.

\subsection{When Na\"ive Transfer Fails}
\label{sec:naive_transfer}
If we know in advance that the source and target task is similar enough, transferring the knowledge of the source task is relatively easy.
For, example, one intuitive and na\"ive method, in this case, is to sample a solution near the solution with good performance in the source task.
Alternatively, if the CMA-ES is performed for the source task, we can reuse the final MGD obtained on the source task as the initial MGD for the target task.
The assumption that the tasks are similar is reasonable in practical cases~\cite{vanschoren2019meta}.
However, it is difficult to guarantee it before performing optimization.
Therefore, it is desirable to alleviate dramatic performance degradation even when these tasks are not very similar.
To confirm the robustness of our proposed warm starting method in such situations, we compare the behavior of the proposed method with the following na\"ive transferring methods:
\begin{itemize}
    \item {\it ReuseGMM} : 
    This method samples solutions from the GMM which represents a promising distribution estimated on the source task; that is, the solutions are sampled from the distribution defined in Eq. (\ref{eq:gmm}) throughout the optimization.
    \item {\it ReuseNormal} : This method uses the final mean and covariance matrix obtained on the source task as the initial MGD on the target task. This method is the same as the (WS-)CMA-ES except for the initialization of MGD.
\end{itemize}

Random search is used as the optimization of a source task for all methods except for ReuseNormal; in ReuseNormal, the result of the CMA-ES is used as the source task.
We consider the sphere function and the rotated ellipsoid function defined in Section 4.1; the experimental settings remain the same.

In addition to these transferring methods, we experiment with the CMA-ES and the sep-CMA-ES, which are not transferring methods, as references.
When the offset changes largely between the source and target tasks, these non-transferring methods become advantageous, as shown in Section \ref{sec:exp_gamma_similarity}.

\begin{figure}[t]
\vspace{-6mm}
    \centering
    \hspace*{\fill}
    \subfloat[][Sphere Function \label{subfig:best_mean_diff}]{
        \includegraphics[width=0.44\linewidth]{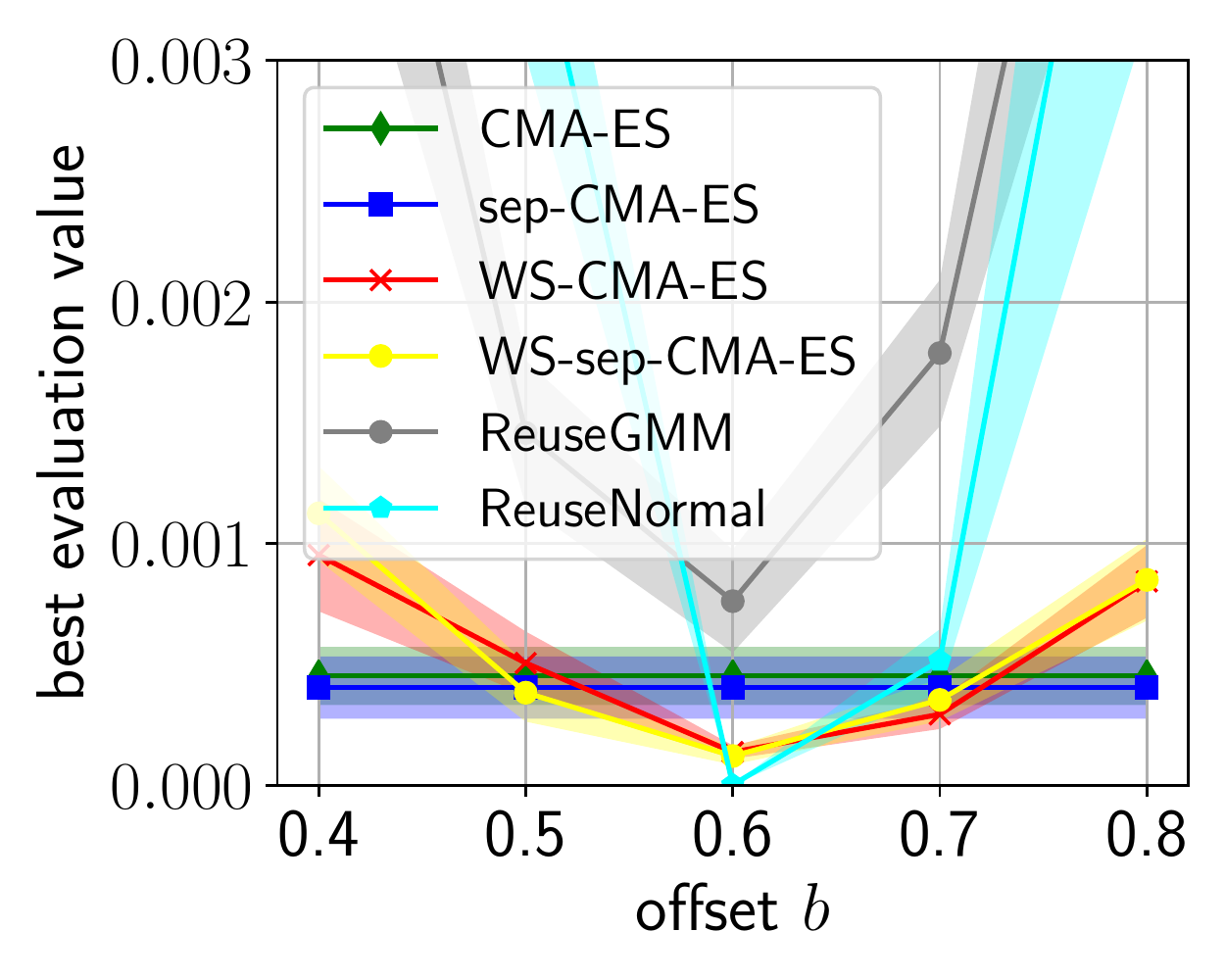}
    }
    \hspace*{\fill}
    \subfloat[][Rotated Ellipsoid Function \label{subfig:exp_gamma_similarity}]{
        \includegraphics[width=0.44\linewidth]{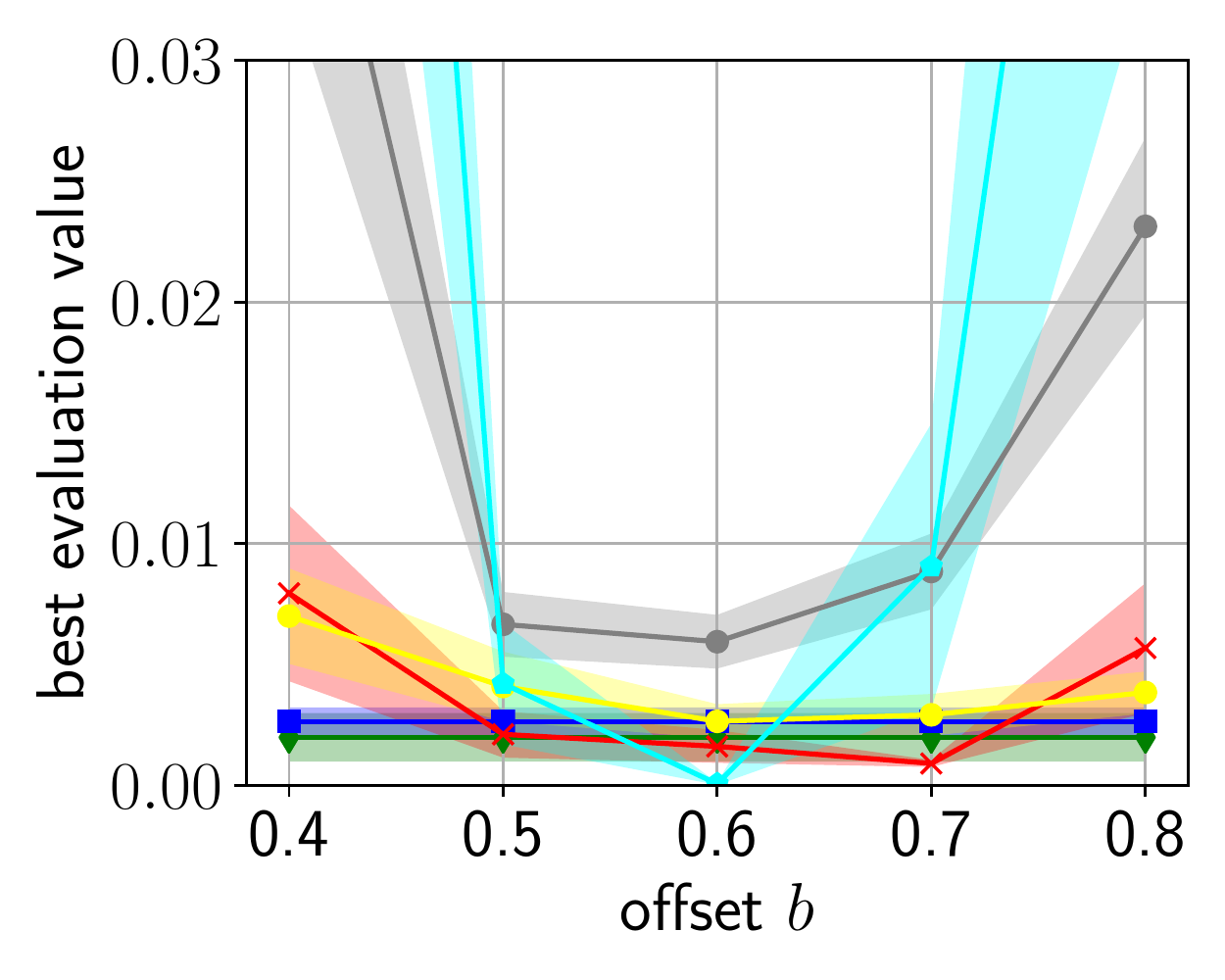}
    }
    \hspace*{\fill}
    \caption{Comparing the proposed methods with naive transferring methods. The mean (line) and the standard error (shadow) over $20$ runs are shown. The results of the CMA-ES and the sep-CMA-ES, which are not transferring methods, are included as references.}
    \label{fig:exp_naive_transfer_comparison}
\end{figure}

Figure \ref{fig:exp_naive_transfer_comparison} presents the results of the experiments over 20 runs.
As expected, ReuseNormal shows the best performance on offset $b = 0.6$ where the source and target tasks are the same.
However, the performance of ReuseNormal deteriorates drastically when the offset is changed.
This is because ReuseNormal converges more than necessary near the optimal solution of the source task even when the optimal solution of the target task is largely different.
In this case, it takes significant time to move from the promising region estimated by the source task, which impairs the performance of ReuseNormal.
In contrast, the proposed methods, WS-CMA-ES and WS-sep-CMA-ES, are less dependent on how long the optimization is performed on the source task, which leads to relatively less performance degradation even in such cases.
Similar to the case of ReuseNormal, ReuseGMM, which does not adapt during optimization, is strongly affected by the dissimilarity of the tasks.
This demonstrates the necessity of the adaptation toward the optimal solution direction of the target task by the CMA-ES (or sep-CMA-ES).
In conclusion, compared with the na\"ive transferring methods, which strongly assume that the tasks are similar, the proposed method is more robust and efficient to the difference between the source and target tasks.

%% file: manuscripts/5_Experiments_HPO.tex
\section{Experiments for Hyperparameter Optimization}
\label{sec:exp_hpo}

\begin{figure*}[t]
\vspace{-4mm}
    \centering
    \hspace*{\fill}
    \subfloat[][HPO of LightGBM on full Toxic Challenge data. \label{subfig:lgbm10per}]{
        \includegraphics[width=0.45\linewidth]{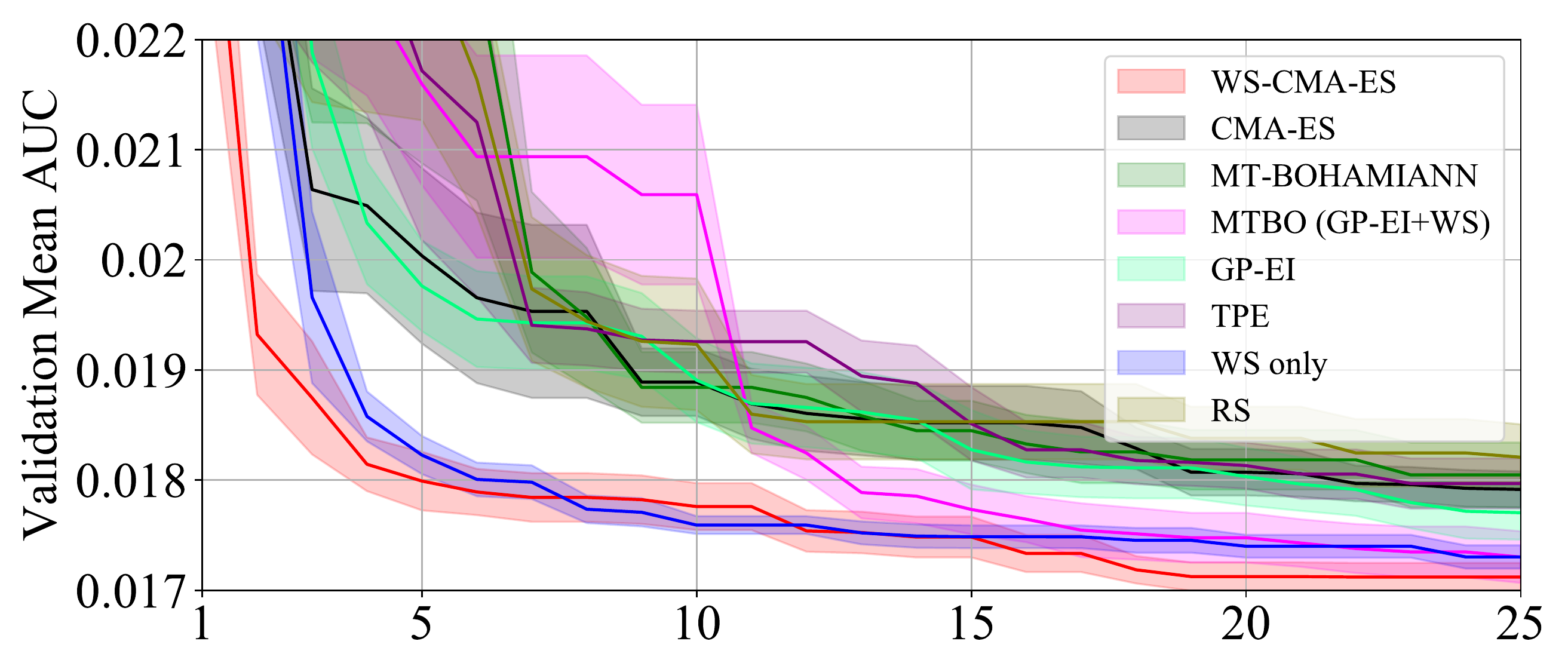}
    }
    \hspace*{\fill}
    \subfloat[][HPO of MLPs on full MNIST. \label{subfig:mnist10per}]{
        \includegraphics[width=0.45\linewidth]{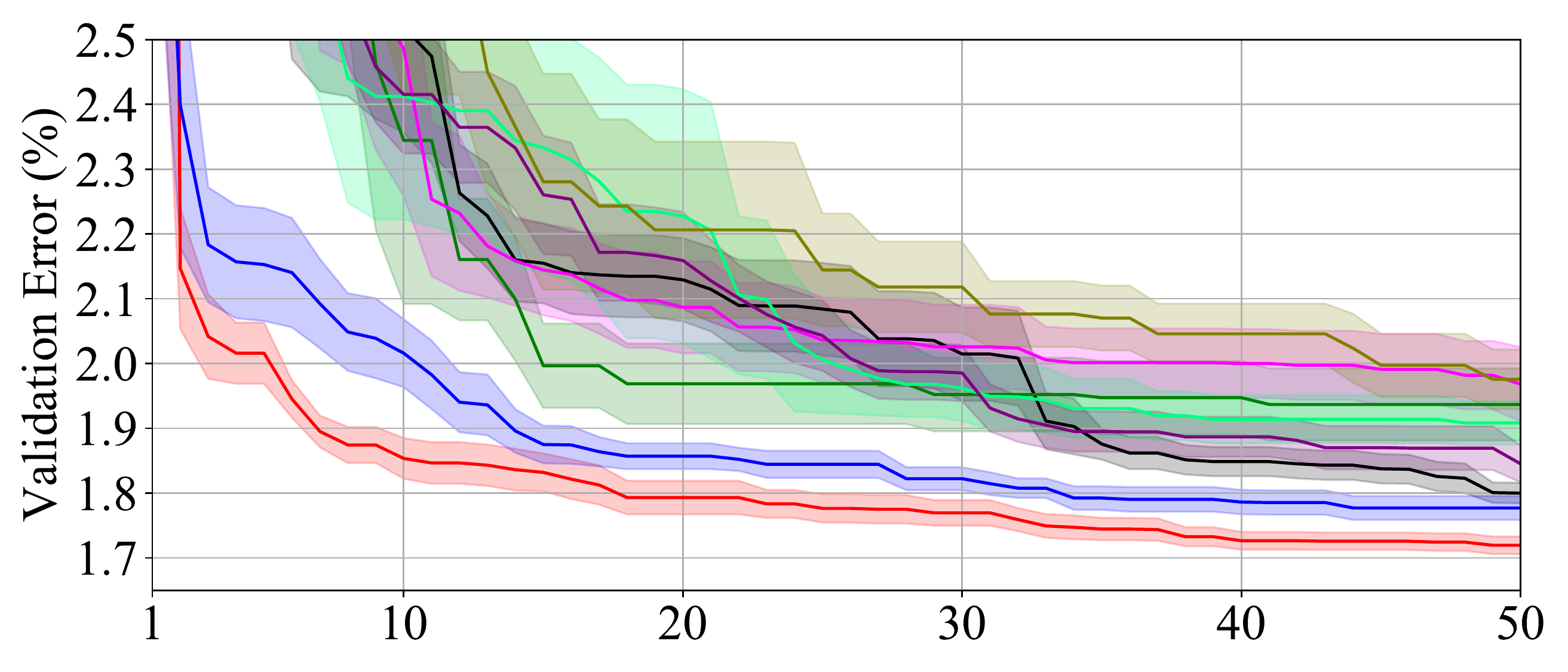}
    }    
    \hspace*{\fill}
    \vspace{-4mm}
    \\
    \hspace*{\fill}
    \subfloat[][HPO of MLPs on full Fashion-MNIST.\label{subfig:f-mnist10per}]{
        \includegraphics[width=0.45\linewidth]{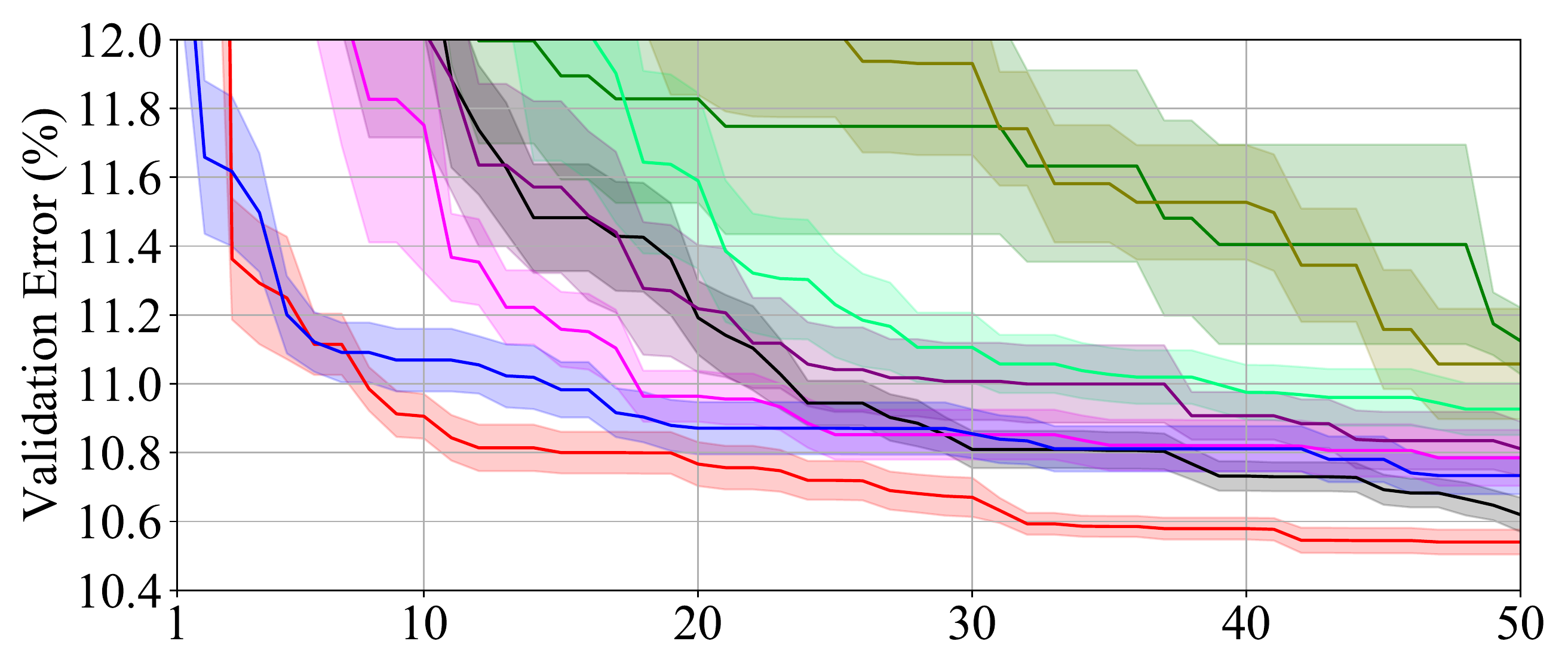}
    }
    \hspace*{\fill}
    \subfloat[][HPO of CNNs on full CIFAR-100.\label{subfig:cnn10per}]{
        \includegraphics[width=0.45\linewidth]{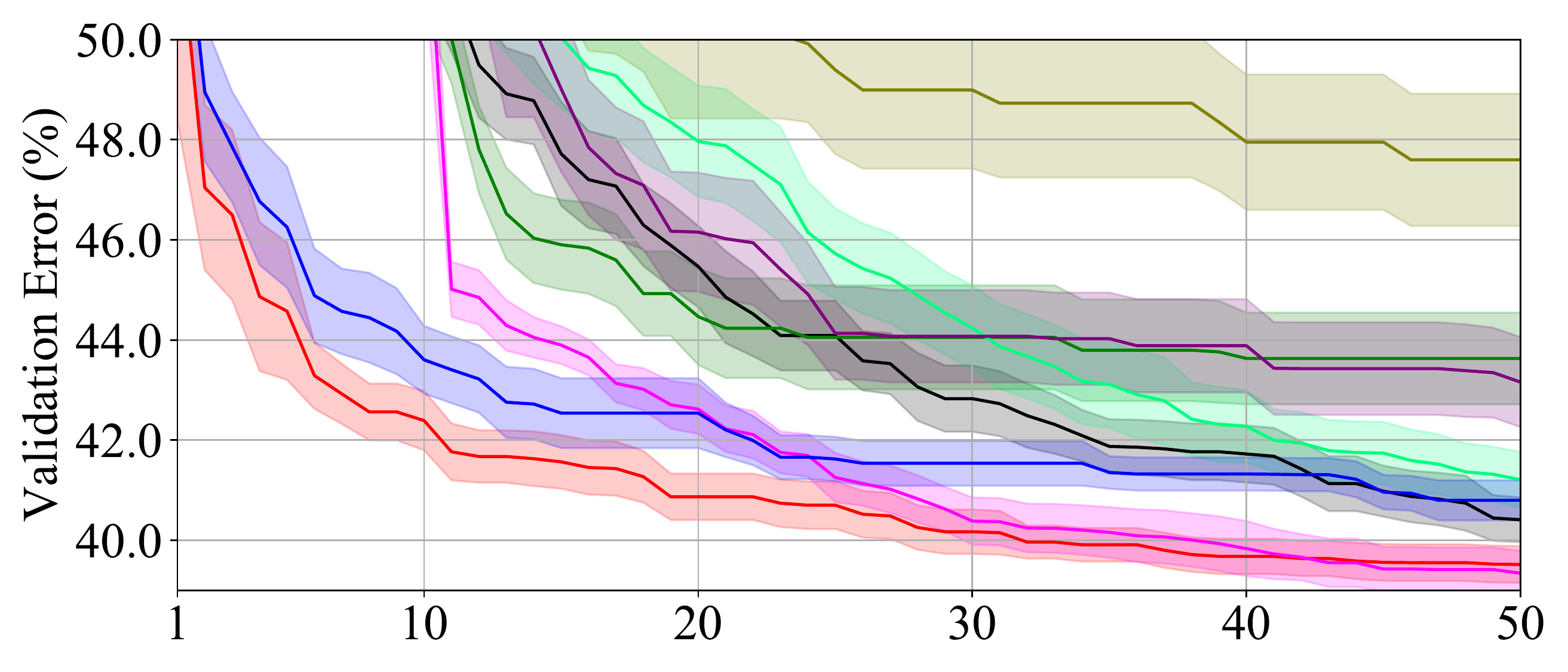}
    }
    \hspace*{\fill}
    \caption{Experiments with warm starting optimization using a result of the HPO for a subset of each dataset. 
    Warm starting methods used a result of the HPO on 1/$10^{\rm th}$ of each dataset as prior knowledge.
    The horizontal axis represents the number of evaluations.
    We plotted the mean and the standard error of the best evaluation value over 12 runs.
    }
    \label{fig:mlp-subsampling-exp}
\end{figure*}


\begin{figure}[t]
\vspace{-4mm}
    \centering
    \hspace*{\fill}
    \subfloat[][MLPs trained on full MNIST]{
        \includegraphics[width=0.45\linewidth]{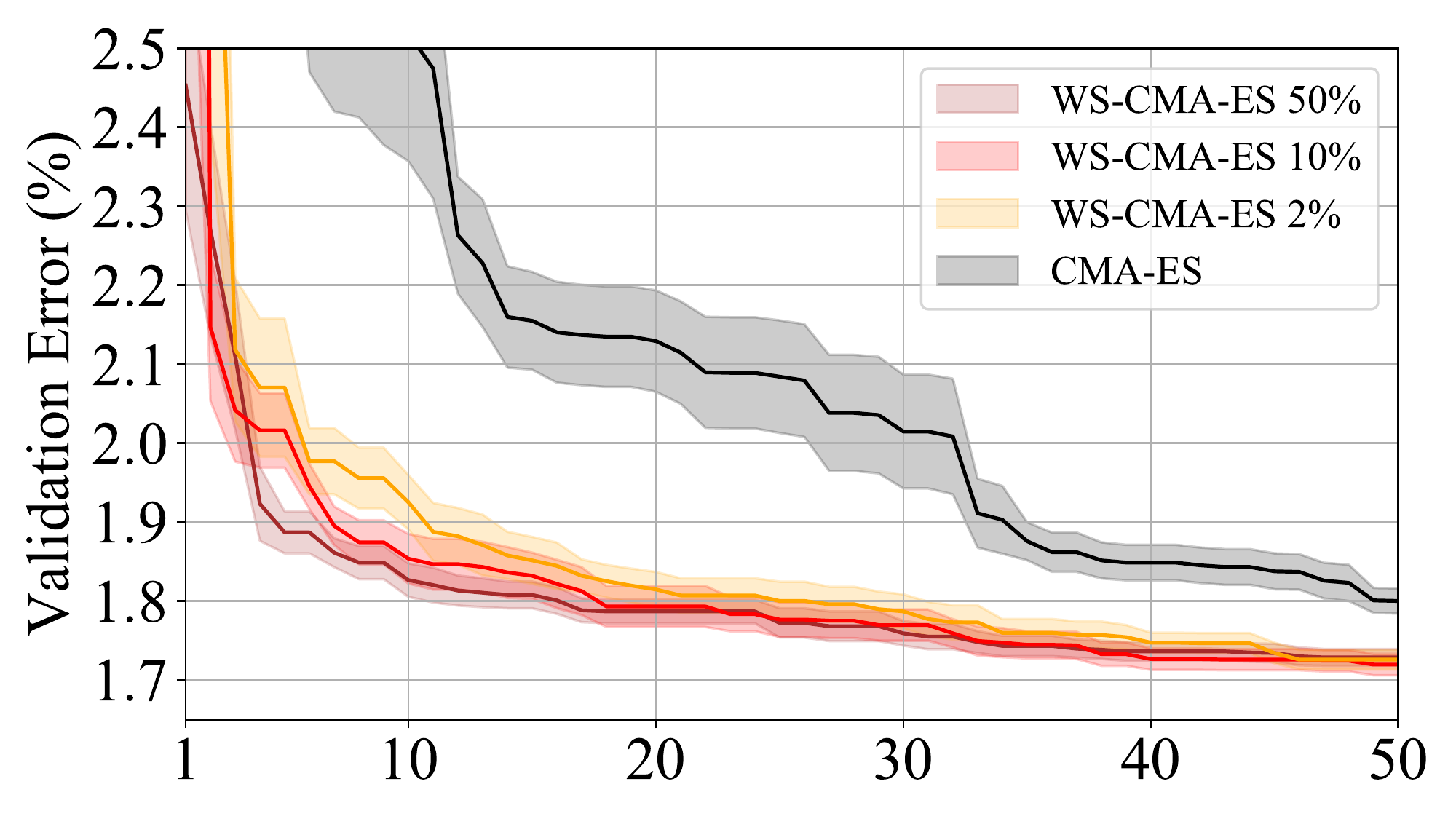}
    }
    \hspace*{\fill}
    \subfloat[][MLPs trained on full Fashion-MNIST \label{subfig:comparison-between-ratio}]{
        \includegraphics[width=0.45\linewidth]{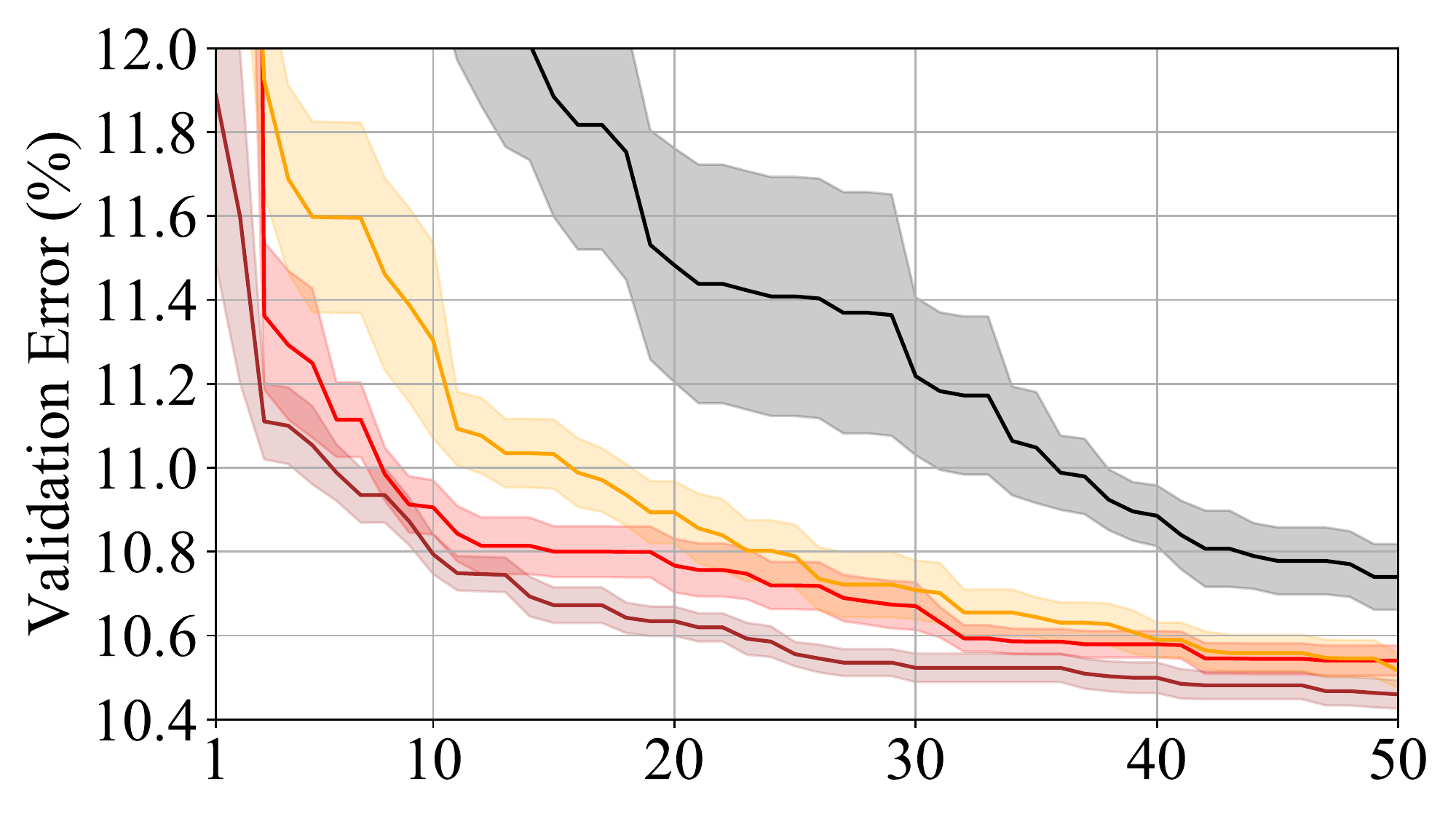}
    }
    \hspace*{\fill}
    \caption{
    Relationship between task similarity and the performance of WS-CMA-ES over 12 times.
    As the size of the dataset approaches that of the complete dataset, the WS-CMA-ES attains faster convergence.
    }
    \label{fig:mlp-subsampling-compare-exp}
\end{figure}

We applied WS-CMA-ES to several HPO problems to verify its effectiveness on HPO.
The experiments comprise the following two practical scenarios:
\begin{itemize}
    \item Warm starting using a result of HPO for a subset of a dataset (Section~\ref{sec:exp_hpo_subset}), and
    \item Warm starting using a result of HPO for another dataset (Section~\ref{sec:exp_hpo_other}).
\end{itemize}
As the baseline methods, we select the
(1) CMA-ES, 
(2) random search (RS)~\cite{bergstra2012random}, 
(3) random sampling from the initial MGD used in WS-CMA-ES (WS-only), 
(4) GP-EI~\cite{NIPS2012_4522}, 
(5) TPE~\cite{bergstra2011algorithms},
(6) MTBO~\cite{swersky2013mtbo}, and 
(7) MT-BOHAMIANN~\cite{springenberg2016bayesian}.
MTBO, which is an extension of GP-EI, and MT-BOHAMIANN are warm starting methods for BO.
TPE is known to provide strong performance in HPO.
Note that we do not use WS-sep-CMA-ES because the performance is similar to WS-CMA-ES in the severely limited budget setting, which is confirmed in Section \ref{sec:naive_transfer}.
We evaluated $100$ hyperparameter settings by RS as prior knowledge in all the experiments to allow every method to transfer the same data fairly.
Each optimization was run $12$ times.
Details of the experimental settings are shown in Appendix A.

\subsection{Warm Starting using a Result of a Subset}
\label{sec:exp_hpo_subset}
We evaluated hyperparameter settings of each machine learning algorithm trained on 10\% of a full dataset.
This result was considered as the source task and was used by the warm starting methods.

\subsubsection{LightGBM on Multilabel Classification}
LightGBM~\cite{ke2017lightgbm} is used as an ML model.
Six hyperparameters shown in Appendix C.1 were optimized in the experiments.
We used the Toxic Comment Classification Challenge data\footnote{https://www.kaggle.com/c/jigsaw-toxic-comment-classification-challenge} as a dataset.
As a metric in the experiments, the mean column-wise area under the receiving operating characteristic curve (ROC AUC) was used.
Note that this measurement is better when the value is higher, so we used $1$ -- AUC as the objective function.

\subsubsection{MLP on MNIST and Fashion-MNIST}
\label{sec:exp_mlp_sub}
The proposed method was applied to the HPO of multilayer perceptrons (MLPs).
We used the MNIST handwritten digits dataset~\cite{lecun1998gradient} and the Fashion-MNIST clothing articles dataset~\cite{xiao2017fashion}.
We optimized eight hyperparameters as shown in Appendix C.2.

\subsubsection{CNN on CIFAR-100}
\label{sec:exp_cnn_sub}
We further applied the proposed method to more sophisticated 8-layer convolutional neural networks (CNNs).
The CNNs were trained on the CIFAR-100 dataset~\cite{krizhevsky2009learning} and have ten types of hyperparameters as described in Appendix C.3.

\subsubsection{Results and Discussion on Knowledge Transfer of a Subset}
Figure~\ref{fig:mlp-subsampling-exp} shows the experiment results.
In each experiment, the proposed method and the WS-only identified better objective metrics much faster than the CMA-ES did.
Further, we found that MTBO yielded better solutions quickly than GP-EI.
Clearly, there was high task similarity between the given tasks that could be exploited by warm starting methods.
WS-CMA-ES and WS-only found better hyperparameter settings in the earlier stage of the optimizations than the others.
In the later stage of the optimization, WS-CMA-ES adapted successfully and converged to better solutions than that of WS-only.
Figure~\ref{fig:mlp-subsampling-exp} (a) shows that WS-CMA-ES and WS-only behave similarly.
This is because the evaluation budget is quite limited, and the update of MGD only happens a few times. 

To observe the correlation between the performance of the WS-CMA-ES and a subset ratio, we applied WS-CMA-ES using prior knowledge of MNIST and Fashion-MNIST of different subset ratios $2\%, 10\%, {\rm and\ } 50\%$.
Figure~\ref{fig:mlp-subsampling-compare-exp} shows the results of the experiments.
We observe that a higher subset ratio tends to result in faster convergence.
The results imply that the sets of relatively good hyperparameters in the source tasks are spatially closer to those in the target tasks, while the best values may not be close, as is claimed in \cite{swersky2013mtbo}.



\begin{figure*}[t]
\vspace{-6mm}
\centering
    \hspace*{\fill}
    \subfloat[][MLPs trained on Fashion-MNIST. As prior knowledge, the result of HPO of MLPs trained on MMIST was used. \label{subfig:mnist50per}]{
        \includegraphics[width=0.45\linewidth]{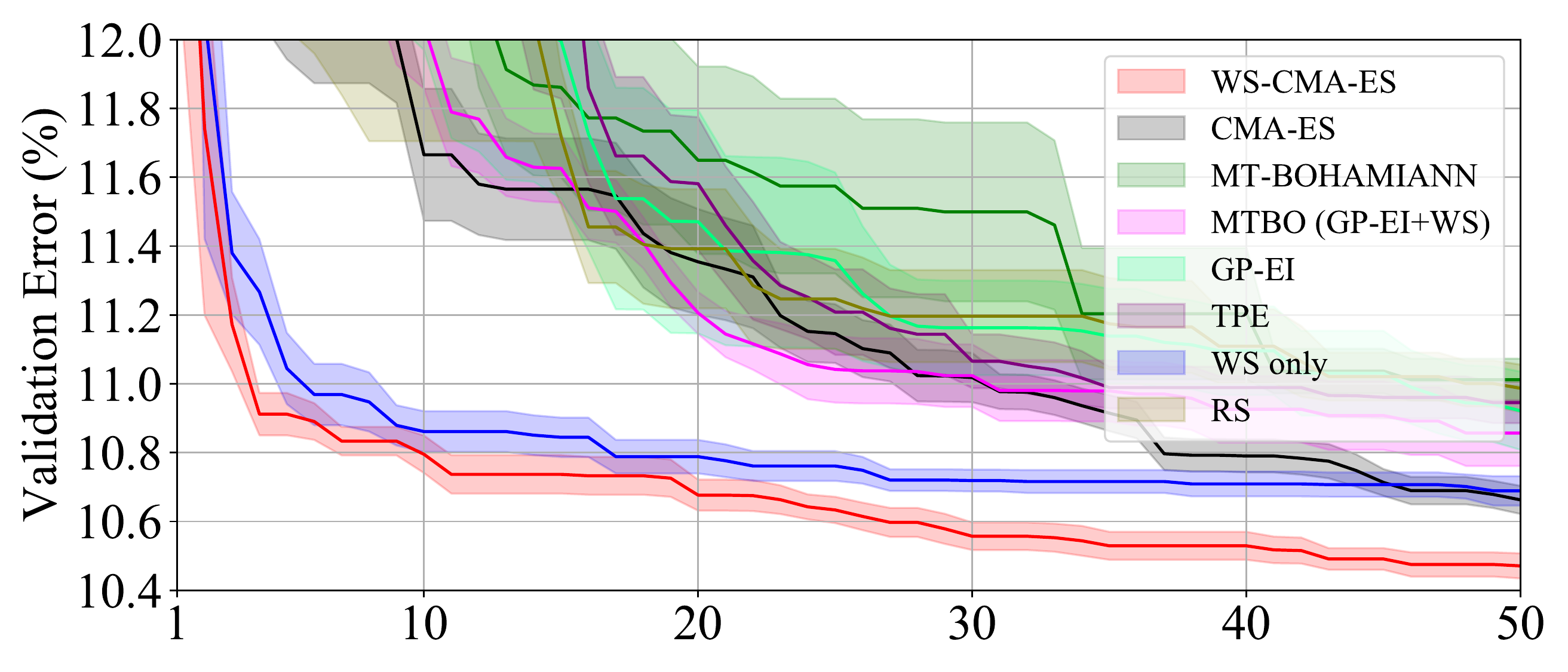}
    }
    \hspace*{\fill}
    \subfloat[][CNNs trained on CIFAR-10. As prior knowledge, the result of HPO of CNNs trained on SVHN was used.\label{subfig:f-mnist50per}]{
        \includegraphics[width=0.45\linewidth]{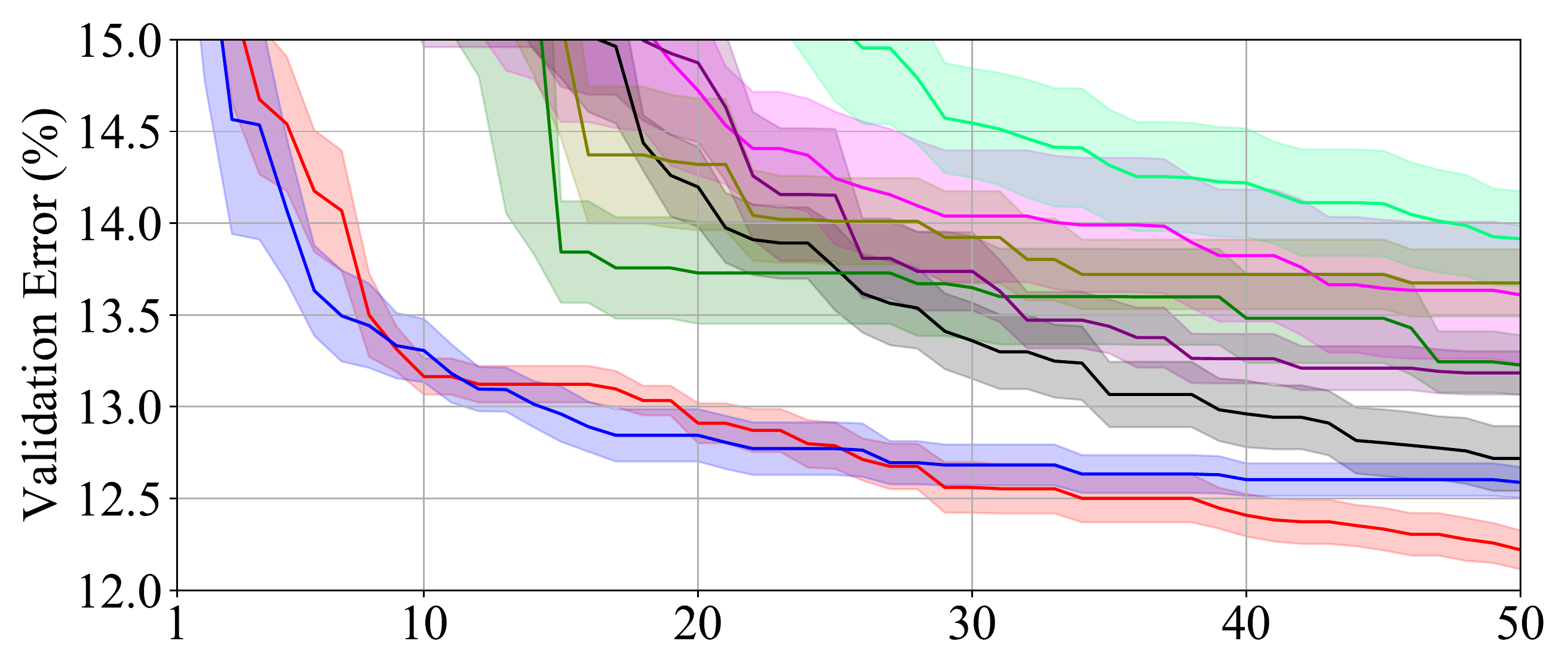}
    }
    \hspace*{\fill}
    \vspace{-1mm}
    \caption{Experiments with warm starting optimizations using the result of HPO on another dataset.}
    \label{fig:mlp-another-dataset-exp}
    \vspace{-3mm}
\end{figure*}

\subsection{Warm Starting using a Result of Another Dataset}
\label{sec:exp_hpo_other}
This section examines what happens when prior knowledge of a different dataset is utilized by warm starting methods.
We carried out experiments to demonstrate the effectiveness of the proposed method in such a practical situation.

\subsubsection{Using the Knowledge of MLP on MNIST for MLP on Fashion-MNIST}
\label{sec:another_fashion_from_mnist}
We first trained MLPs on MNIST and then transferred the result to the HPO of Fashion-MNIST.
The architecture of the MLPs and their hyperparameters are the same as those described in Section~\ref{sec:exp_mlp_sub}.

\subsubsection{Using the Knowledge of CNN on SVHN for CNN on CIFAR-10}
\label{sec:another_cifar_from_svhn}
We optimized the 8-layer CNNs.
Hyperparameters for this model are the same as those of the model optimized earlier (see Section~\ref{sec:exp_cnn_sub}).
CNNs initially learned the Street View House Numbers (SVHN) dataset~\cite{netzer2011reading}.
Next, the warm starting methods employed the knowledge to obtain the optimal hyperparameter settings for CNNs trained on CIFAR-10.

\subsubsection{Results and Discussion on the Knowledge Transfer of Another dataset}
Figure~\ref{fig:mlp-another-dataset-exp} shows the results of the experiments.
The proposed method exhibited outstanding convergence speed in the experiments and found better hyperparameter settings far more quickly than the CMA-ES.
Although MTBO also successfully found better solutions than GP-EI, the performance of MTBO was not considerably better than that of RS.
In fact, MTBO required approximately 25 evaluations to find better hyperparameter settings than GP-EI in the results described in Figure~\ref{fig:mlp-subsampling-exp} (b), (d).
According to Figure~\ref{fig:mlp-another-dataset-exp}, however, it required approximately 40 evaluations in these experiments.
Contrarily, the WS-CMA-ES identified better hyperparameter settings than the CMA-ES in approximately 25 and 30 evaluations in the experiments using a small dataset and experiments using another dataset, respectively.
This is probably because knowledge transfer from other datasets is more difficult than knowledge transfer from a subset of a dataset.
MTBO obtains promising solutions using the approximation of the entire search space, but the WS-CMA-ES obtains promising solutions using that of only the promising region.
The former approximation requires more observations to yield promising solutions compared with the latter.
This may be the reason for the effectiveness of WS-CMA-ES in knowledge transfer from another dataset.
This behavior can also be confirmed with the transfer HPO experiments with other datasets, which are provided in Appendix \ref{supple_sec:another}.

%% file: manuscripts/6_Related_Work_Discussion.tex
\section{Related Work and Discussion}
Various types of warm starting methods for BO have been actively studied in the HPO context.
These methods model the relationship between tasks using a variety of ways, such as a Gaussian process~\cite{swersky2013mtbo,poloczek2016warm,poloczek2017multi,wistuba2018scalable,feurer2018scalable}, deep neural networks~\cite{springenberg2016bayesian,kim2017learning}, and Bayesian linear regression~\cite{perrone2018scalable}.
However, the CMA-ES, which shows outstanding performance in BBO, has not been thoroughly considered in HPO.

One difference between our method and the warm starting methods for BO is in the usage of the source tasks' result.
Most warm starting methods for BO repeatedly construct a probabilistic model using prior knowledge in each iteration.
In contrast, WS-CMA-ES uses prior knowledge only at the inception of optimization.
Therefore, the computational complexity of WS-CMA-ES does not depend on the number of observations.
This enables users to implement the method even when numerous results are available.
Although the meta-feature based warm starting~\cite{feurer2015initializing} can alleviate this computational problem, it is not always possible to prepare such meta-feature for the dataset.
The method of initializing the search space using the result of the source task does not incur extra computational complexity and can be used without such a meta feature~\cite{wistuba2015hyperparameter,perrone2019learning}.

Another difference is that it is usually challenging for most BO approaches to handle the scale variation of objective functions across tasks.
This situation often appears when exploiting prior knowledge in transfer HPO;
for example, the validation error may significantly change across different datasets.
This situation also appears when transferring between different objectives, such as transferring between the result of misclassification error and that of cross entropy.
Salinas et al. introduced a sophisticated semi-parametric approach to deal with such a heterogeneous scale~\cite{salinas2019copula}.

%% file: manuscripts/7_Conclusion.tex
\section{Conclusion and Future Work}
\label{sec:conclusion}
We proposed the WS-CMA-ES, a simple and effective warm starting strategy for the CMA-ES.
The proposed method was designed based on the theoretical definitions of a promising distribution and task similarity.
It initializes MGD in the CMA-ES by approximating the promising distribution on a source task.
This knowledge transfer performs well especially when a target task is similar to a source task in terms of the defined task similarity, which is confirmed by our experiments.
Experiments with synthetic and HPO problems confirm that WS-CMA-ES is effective, even with low budgets or when the source and target tasks are not very similar.


The main limitation of this study is the assumption of task similarity.
From our experiments and the desirable results of warm starting methods that assume task similarity (e.g., \cite{bardenet2013collaborative,yogatama2014efficient}), we hypothesize that HPO tasks are often similar as long as so are they intuitively.
However, WS-CMA-ES can be worse than the CMA-ES when the similarity between the source and target tasks is low, as shown in \Cref{fig:exp_performance_vs_similarity}.
Automatic detection of task dissimilarity and switching back to the original CMA-ES is essential for this method to be more convincing and reliable.

%% file: manuscripts/8_Appendix.tex
\appendix

\section{Details of Experimental Setups}
\label{sec:exp_setting}
We used pycma\footnote{https://github.com/CMA-ES/pycma} library to obtain the results of the CMA-ES.
Note that pycma is also used to implement WS-CMA-ES, as WS-CMA-ES is the same as the CMA-ES except for warm starting.
We used BoTorch (used MultiTaskGP class to implement MTBO)\footnote{https://github.com/pytorch/botorch} for GP-EI~\cite{NIPS2012_4522} and MTBO~\cite{swersky2013mtbo}.
MT-BOHAMIANN~\cite{springenberg2016bayesian} and TPE~\cite{bergstra2011algorithms} were transplanted from RoBO\footnote{https://github.com/automl/RoBO} and Hyperopt\footnote{https://github.com/hyperopt/hyperopt} for each method.

All the experiments were conducted on AI Bridging Cloud Infrastructure (ABCI)\footnote{https://abci.ai/}.
The GPU was NVIDIA Tesla V100 (SXM2), and the CPU was Intel Xeon Gold 6148 (27.5M Cache, 2.40GHz, 40core).
It required about 6 days to complete an optimization of one warm starting setting.\footnote{https://abci.ai/en/about\_abci/computing\_resource.html}

The CMA-ES has two control parameters, i.e. sample size $\lambda$ and an initial distribution.
We set $\lambda = 8$ as a sample size.
In addition, we set $\mathcal{N}(0.5, 0.2^2)$ as the initial distribution for each variable that is bounded in $[0, 1]$ and this setting is identical to the setting described in \cite{loshchilov2016cma}.
We followed the default settings of the pycma library about the other control parameters of the CMA-ES.
In MTBO, MT-BOHAMIANN, GP-EI, and TPE, the number of initial samples was set to $10$.
In all the experiments, $\alpha$ and $\gamma$ in WS-CMA-ES were set to $0.1$ for each.
Note that users do not have to change $\alpha$ as long as the bound for each variable is $[0, 1]$.
We confirmed the robustness of WS-CMA-ES for the variation of $\alpha$ value and $\gamma$ value in Section~\ref{supple_sec:hp_ws_cma}.

\section{Details of Synthetic Problems}

\begin{table*}[t]
\vspace{+2mm}
    \centering
    \hspace*{\fill}
    \subfloat[][The sphere function. \label{subfig:bench_sphere}]{
        \begin{tabularx}{0.49\textwidth}{p{9em}p{5em}p{6em}}
                \toprule
                Method & $\ \,b$ & $ \ \ \ \ \ \ \ \ \ \ f$  \\ 
                \midrule
                CMA-ES    & $\ -$     & $0.43  \ \ \pm 0.1$ \\
                WS-CMA-ES & $0.4$ & $1.3 \ \ \ \ \pm 0.3$ \\
                            & $0.5$ & $0.26  \ \ \pm 0.06$ \\
                            & $0.6$ & $0.073 \pm 0.02$ \\
                            & $0.7$ & $0.27  \ \ \pm 0.07$ \\
                            & $0.8$ & $0.82  \ \ \pm 0.2$ \\
                 \bottomrule
        \end{tabularx}
    }    
    \hspace*{\fill}
    \subfloat[][The rotated ellipsoid function.\label{subfig:bench_rotell}]{
        \begin{tabularx}{0.49\textwidth}{p{9em}p{5em}p{6em}}
            \toprule
            Method & $\ \,b$ & $ \ \ \ \ \ \ \ \ f$  \\ 
            \midrule
            CMA-ES    & $\ -$     & $0.25 \pm 0.1$ \\
            WS-CMA-ES & $0.4$ & $0.26 \pm 0.06$ \\
                        & $0.5$ & $0.21 \pm 0.06$ \\
                        & $0.6$ & $0.14 \pm 0.05$ \\
                        & $0.7$ & $0.15 \pm 0.06$ \\
                        & $0.8$ & $0.38 \pm 0.2$ \\
             \bottomrule
        \end{tabularx}
    }
    \hspace*{\fill}
    \caption{Details of the result of each benchmark function. 
    $f$ is the mean ($\pm$ the standard error) of the best evaluation value, and $b$ is a coefficient for a prior task.
    We magnify the original values by $10^3$ times for the sphere function and $10^2$ times for the rotated ellipsoid function.
    }
    \label{tab:bench-exp}
    \vspace{0mm}
\end{table*}

The optimizations for each problem were run $20$ times.
We set the number of evaluations to $50$ for both problems.
Both $\gamma_1$ and $\gamma_2$ for calculating $\gamma$-similarity were set to $0.1$.
We used the Monte Carlo method to calculate $\gamma$-similarity.

Table~\ref{tab:bench-exp} shows the results of the sphere function and the rotated ellipsoid function.
The column $f$ shows the mean and the standard error of the best evaluation value for each experimental setting.
Note that $b$ is a coefficient for a prior task that WS-CMA-ES used for warm starting.

\section{Experimental Settings for Hyperparameter Optimization}
In all the experiments, each dataset was split into 80\% training data and 20\% validation data.
A subset of each dataset was yielded by specifying the indices of data up to $\lfloor S \times |\mathcal{D}_{\rm train}|\rfloor$ where $S$ is a subset ratio, and $|\mathcal{D}_{\rm train}|$ is the size of a training data.
For example, if $S=0.1$ and the size of the complete dataset is $50,000$, we used training data of size $4,000$ and validation data of size $10,000$.
We used the above-mentioned training data (or the subset) to train the model and the validation data to measure performance.
In a series of experiments, we rounded integer parameters.

\subsection{Details of the Experiments for LightGBM}
\label{supple_sec:lgbm}

We optimized the hyperparameters shown in Table~\ref{tab:hp_lgbm}.
We implemented the preprocessing of the dataset in the identical way\footnote{https://www.kaggle.com/peterhurford/lightgbm-with-select-k-best-on-tfidf}.
The mean column-wise ROC AUC was measured by $5$-fold cross validation.
\begin{table}[t]
\vspace{+2mm}
\caption{Details of the hyperparameters of LightGBM.}
\label{tab:hp_lgbm}
\begin{center}
\begin{tabularx}{0.49\textwidth}{p{9.5em}p{2.5em}p{10em}}
        \toprule
         Hyperparameters & Scale & Range \\ 
         \midrule
         Learning Rate                            & log & $[10^{-3}, 1.0]$\\
         \# of Leaves                                & log & $[8, 128]$\\
         Bagging Fraction                       & linear & $[0.1, 0.9]$ \\
         Feature Fraction                           & linear & $[0.1, 0.9]$ \\
         L1 Regularization                          & log & $[10^{-1}, 10]$ \\
         L2 Regularization                          & log & $[10^{-1}, 10]$ \\
         \bottomrule
\end{tabularx}
\end{center}
\end{table}

\subsection{Details of the Experiments for MLP}
\label{supple_sec:mlp}
We optimized the hyperparameters shown in Table~\ref{tab:hp_mlp}.
As the preprocessing of each dataset (MNIST and Fashion-MNIST), we apply a random crop for training and a central crop for validation.
The size of both the MNIST and Fashion-MNIST dataset is $50,000$.
The architecture of MLP is shown in Table~\ref{tab:hp_mlp}.
We trained for a total of $20$ epochs and used stochastic gradient descent with Nesterov accelerated gradient method~\cite{nesterov1983method}.

\begin{table}[t]
\caption{Details of the hyperparameters of MLP.}
\label{tab:hp_mlp}
\begin{center}
\begin{tabularx}{0.49\textwidth}{p{9.5em}p{2.5em}p{15em}}
        \toprule
         Hyperparameters & Scale & Range \\
         \midrule
         Batch Size                                 & log & $[32, 256]$\\
         Learning Rate                              & log & $[5.0 \cdot 10^{-3}, 5.0 \cdot 10^{-1}]$ \\
         Weight Decay                               & log & $[5.0 \cdot 10^{-6}, 5.0 \cdot 10^{-2}]$ \\
         Momentum                                   & linear & $[0.8, 1.0]$ \\
         \# of Units (${\rm U_1}$)                           & log & $[32, 512]$ \\
         \# of Units (${\rm U_2}$)                           & log & $[32, 512]$ \\
         Dropout Rate (Layer 1)                       & linear & $[0.0, 1.0]$ \\
         Dropout Rate (Layer 2)                       & linear & $[0.0, 1.0]$ \\
         \bottomrule
\end{tabularx}
\end{center}
\end{table}

\subsection{CNN on CIFAR-100}
\label{supple_sec:cnn}
We optimized the hyperparameters shown in Table~\ref{tab:hp_cnn}.
We implemented the preprocessing of each dataset (CIFAR-10, CIFAR-100, and SVHN) in the identical way that is described in \cite{wrn}.
The size of the CIFAR-10 and CIFAR-100 dataset is $50,000$ and that of SVHN is $73,257$.
The architecture of CNN is shown in Table~\ref{tab:arc_cnn}.
We trained for a total of $160$ epochs and used stochastic gradient descent with Nesterov accelerated gradient method.

\begin{table}[t]
\vspace{+2mm}
\caption{Details of the hyperparameters of CNN.
$C_k$, FC, and CL stand for the $k$-th convolutional layer, fully-connected layer, and classification layer, respectively.
}
\label{tab:hp_cnn}
\begin{center}
\begin{tabularx}{0.49\textwidth}{p{8em}p{3em}p{15em}}
         \toprule
         Hyperparameters & Scale & Range \\ 
         \midrule
         Batch Size                                 & log & $[32, 256]$\\
         Learning Rate                              & log & $[5.0 \cdot 10^{-3}, 5.0 \cdot 10^{-1}]$ \\
         Weight Decay                               & log & $[5.0 \cdot 10^{-6}, 5.0 \cdot 10^{-2}]$ \\
         Momentum                                   & linear & $[0.8, 1.0]$ \\
         \# of Kernels (${\rm C_1}$)           & log & $[16, 128]$ \\
         \# of Kernels (${\rm C_2}$)           & log & $[16, 128]$ \\
         \# of Kernels (${\rm C_3}$)           & log & $[16, 128]$ \\
         \# of Kernels (${\rm C_4}$)           & log & $[16, 128]$ \\
         Dropout Rate (FC)                       & linear & $[0.0, 1.0]$ \\
         Dropout Rate (CL)                       & linear & $[0.0, 1.0]$ \\
         \bottomrule
\end{tabularx}
\end{center}
\end{table}

\begin{table*}[t]
\centering
\caption{Details of the architecture of the MLP.}
\vspace{+2mm}
\label{tab:arc_cnn}
\begin{tabularx}{\textwidth}{p{11em}p{7em}p{20em}}
\toprule
Components & Output Size & Layer Type \\
\midrule
Fully-connected Layer 1 & ${\rm U_1}$ & ReLU, dropout \\ 
\midrule
Fully-connected Layer 2 & ${\rm U_2}$ & ReLU, dropout \\ 
\midrule
Classification Layer & $10$ & $10$ dimensional fully-connected, softmax \\ 

\bottomrule

\end{tabularx}
\vspace{0mm}
\end{table*}

\begin{table*}[t]
\centering
\caption{Details of the architecture of the CNN.}
\vspace{+2mm}
\label{tab:arc_cnn}
\begin{tabularx}{\textwidth}{p{11em}p{7em}p{20em}} 
\toprule
Components & Output Size & Layer Type \\
\midrule
Convolution 1 & $32 \times 32 \times {\rm C_1}$ & $5 \times 5 {\rm\ conv} \times {\rm C_1}$, padding 2 \\ 
Max Pooling & $15 \times 15 \times {\rm C_1}$ & $3 \times 3$, stride $2$, ReLU \\ \midrule

Convolution 2 & $15 \times 15 \times {\rm C_2}$ & $5 \times 5 {\rm\ conv} \times {\rm C_2}$, padding $2$, ReLU \\ 
Average Pooling & $7 \times 7 \times {\rm C_2}$ & $3 \times 3$, stride $2$, batch normalization \\ \midrule

Convolution 3 & $7 \times 7 \times {\rm C_3}$ & $5 \times 5 {\rm\ conv} \times {\rm C_3}$, padding $2$, ReLU \\ 
Average Pooling & $3 \times 3 \times {\rm C_3}$ & $3 \times 3$, stride $2$, batch normalization, dropout \\ \midrule

Fully-connected Layer & ${\rm C_4} \times 1$ & ${\rm C_4}$ dimensional fully-connected, dropout\\ 
Classification Layer & $10\ {\rm or}\ 100 \times 1$ & $10$ or $100$ dimensional fully-connected, softmax \\ 

\bottomrule

\end{tabularx}
\vspace{0mm}
\end{table*}

\section{Warm Starting Using a Result of Another Dataset: Extra Results}
\label{supple_sec:another}
In the main paper, we transferred the knowledge of MNIST to the optimization of the MLP trained on Fashion-MNIST.
Likewise, we transferred the knowledge of SVHN to the optimization of the CNN trained on CIFAR-10.
In this supplementary, we provide results of knowledge transfer of different directions.
In other words, we conducted the optimization of MNIST using the prior knowledge of the result of Fashion-MNIST and the optimization of SVHN using the prior knowledge of the result of CIFAR-10.
The main difference of the knowledge transfer between the main paper and this supplementary is the level of difficulty of each dataset.
We herein used the results of Fashion-MNIST and CIFAR-10, which are more difficult to solve by the same model than MNIST and SVHN, respectively, as prior knowledge.

\begin{figure*}[t]
    \centering
    \hspace*{\fill}
    \subfloat[][MLPs trained on MNIST. As prior knowledge, the result of HPO of MLPs trained on Fashion-MMIST was used.]{
        \includegraphics[width=0.45\linewidth]{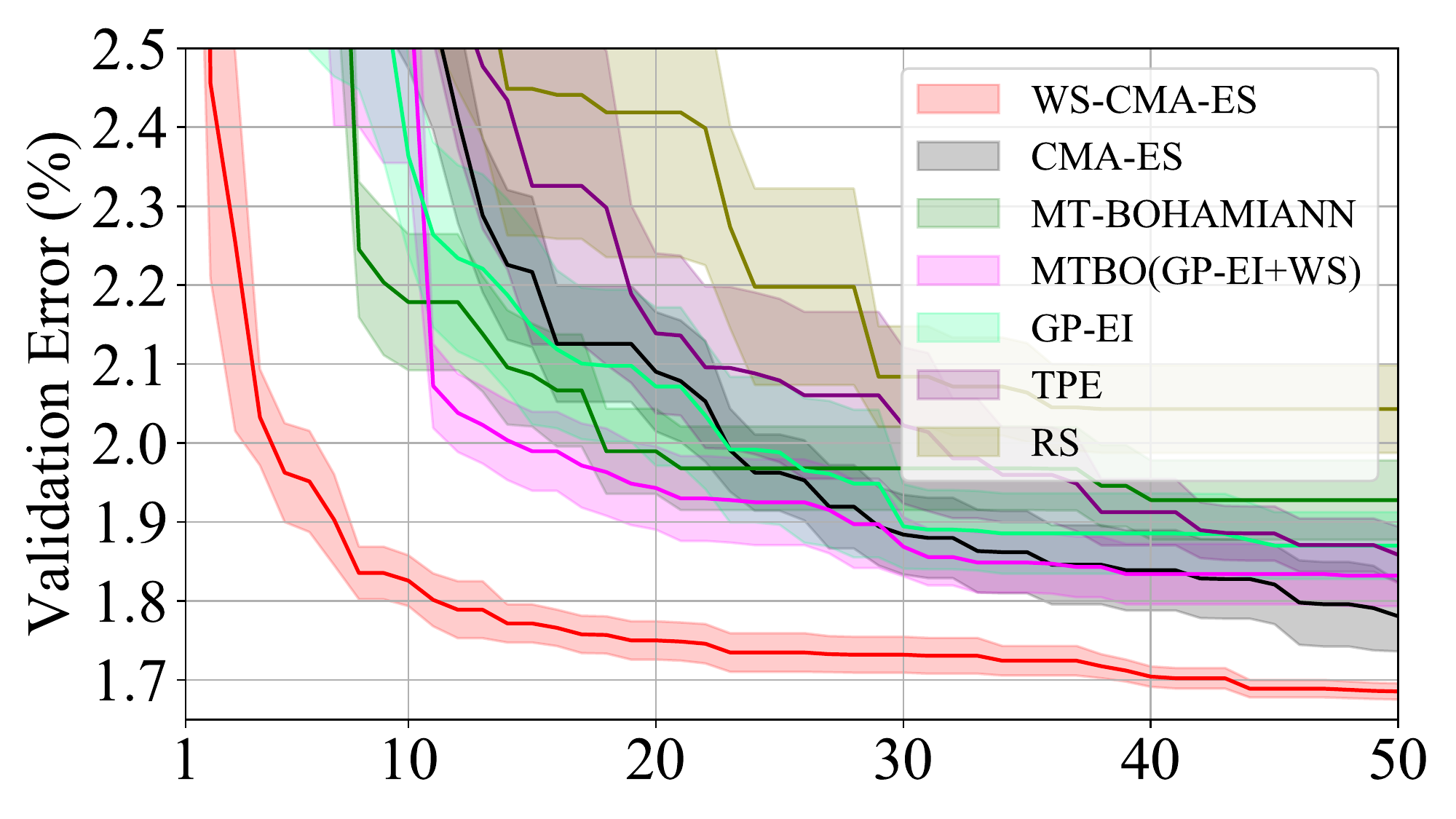}
    }
    \hspace*{\fill}
    \subfloat[][CNNs trained on SVHN. As prior knowledge, the result of HPO of CNNs trained on CIFAR-10 was used.]{
        \includegraphics[width=0.45\linewidth]{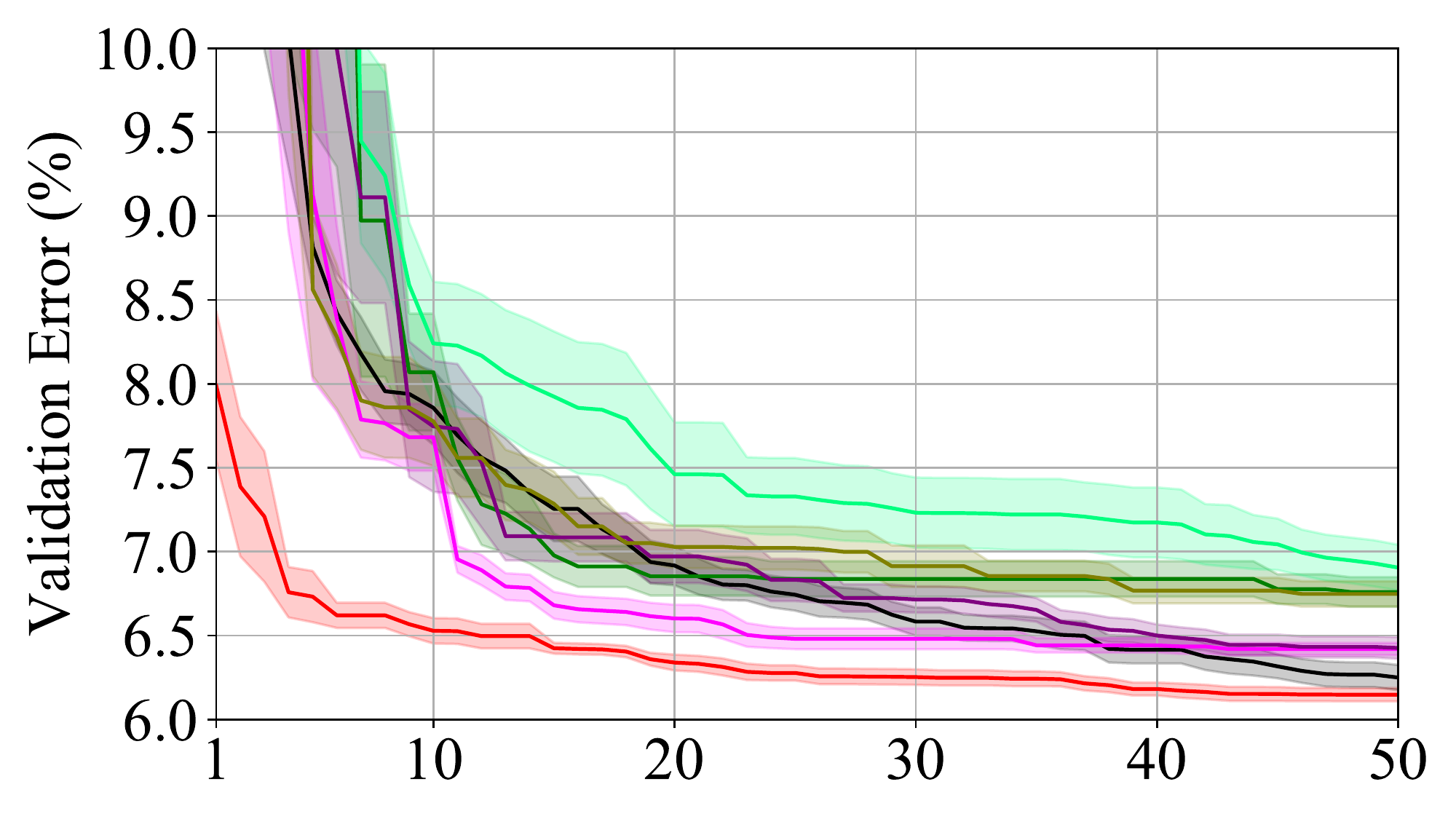}
    }
    \hspace*{\fill}
    \caption{Experiments of warm starting optimizations using the result of HPO on another dataset.
    The directions of knowledge transfer are the opposite of those in the main paper.}
    \label{fig:exp-another-extra}
    \vspace{0mm}
\end{figure*}

Figure~\ref{fig:exp-another-extra} shows the results of the experiments.
In both settings, the proposed method obtained better hyperparameter settings more quickly than the CMA-ES.
Compared to the results shown in Figure~5 (a), (b), the performance of the WS-CMA-ES was not outstanding.
This is probably because the current tasks were easy enough to enable any optimization methods to yield relatively good solutions without any knowledge.
In fact, RS exhibited better performance than GP-EI in Figure~\ref{fig:exp-another-extra} (b).
The proposed method converged faster, even in such a situation.

\section{Characteristics of the Control Parameters of the WS-CMA-ES}
\label{supple_sec:hp_ws_cma}

\subsection{MLPs on MNIST}
To examine the parameter sensitivity of WS-CMA-ES, we implemented WS-CMA-ES with various values of $\alpha$ and $\gamma$.
In the experiments, the WS-CMA-ES optimized the MLPs trained on MNIST that is described in Appendix~\ref{supple_sec:mlp} using prior knowledge on the result of 1/$10^{\rm th}$ of complete MNIST.

First, we conducted experiments of the parameter sensitivity with respect to $\alpha$.
By construction presented in Eq. (3), when the $\alpha$ approaches $0$, a promising distribution for a prior task will be overconfident and overfit to a prior task.
In contrast, as the $\alpha$ goes to $\infty$, a promising distribution for a prior task will take the uncertainty of the objective function into consideration seriously and will be closer to a uniform distribution.
Therefore, we have to balance the trade-off.
In this experiment, we fixed $\gamma=0.1$ and varied $\alpha$ from $0.05$ to $0.25$ in increments of $0.05$.
Note that $\gamma=0.1$ and $\alpha=0.1$ were the default setting of all the experiments provided in the main paper.

\begin{figure*}[t]
    \centering
    \hspace*{\fill}
    \subfloat[][The parameter sensitivity of WS-CMA-ES with respect to $\alpha$]{
        \includegraphics[width=0.45\linewidth]{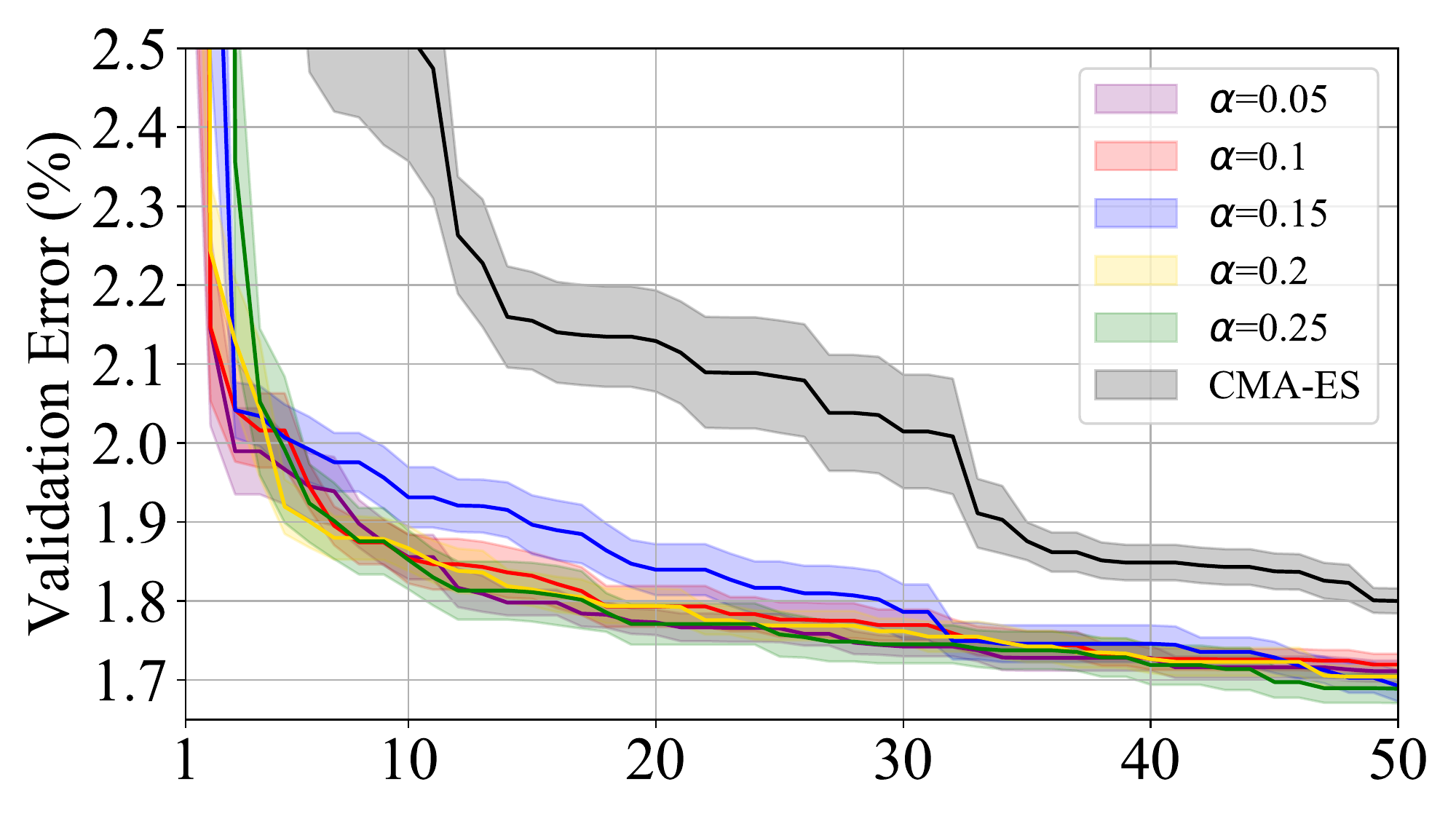}
    }
    \hspace*{\fill}
    \subfloat[][The parameter sensitivity of WS-CMA-ES with respect to $\gamma$ \label{subfig:gamma}]{
        \includegraphics[width=0.45\linewidth]{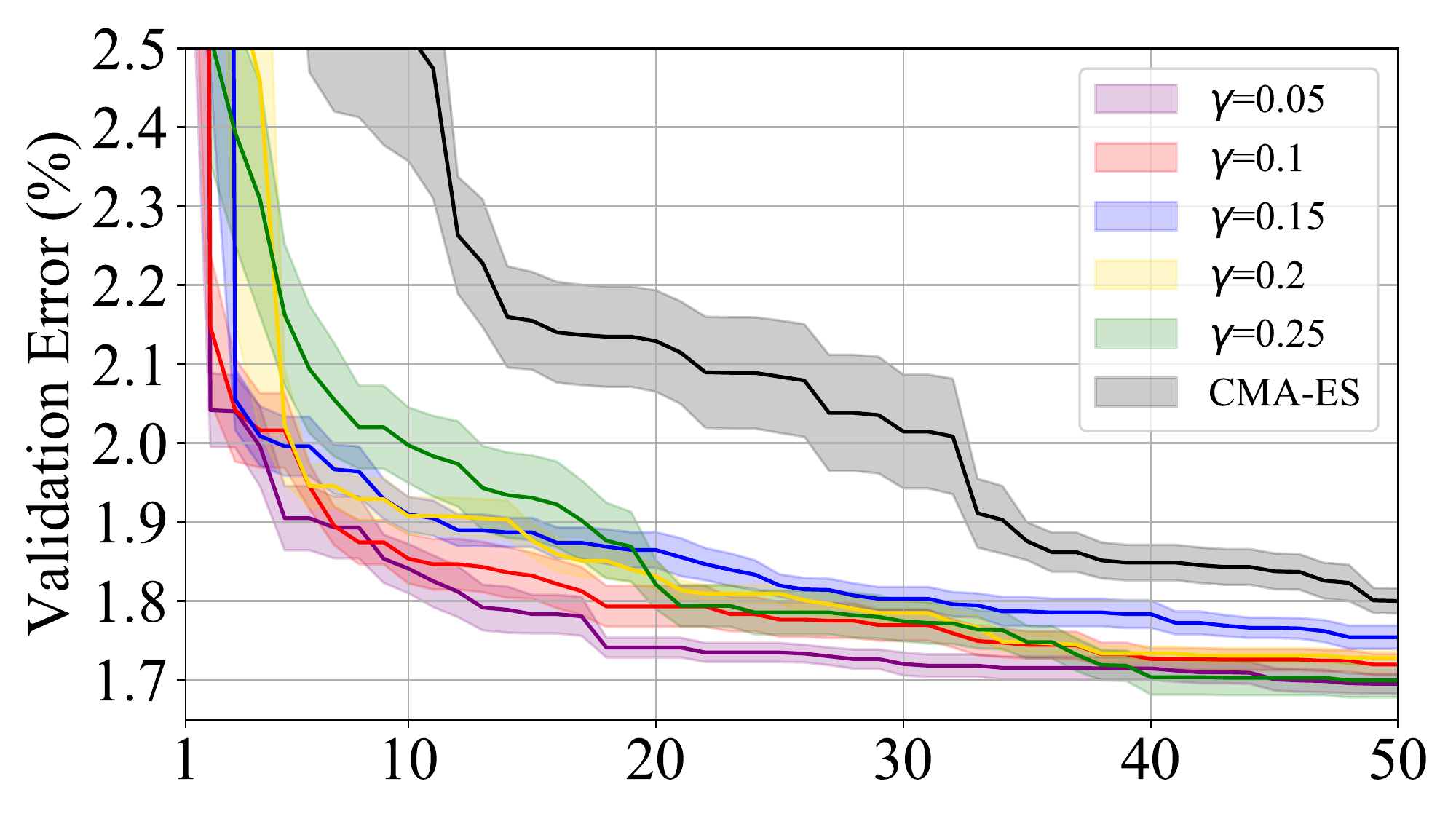}
    }
    \hspace*{\fill}
    \vspace{0mm}
    \caption{Experiments to clarify the parameter sensitivity of the WS-CMA-ES with respect to the control parameters $\alpha$ and $\gamma$ for the MLPs on MNIST.
    $\alpha$ controls the extent of a promising distribution.
    $\gamma$ controls the number of solutions which is used when a promising distribution is constructed.
    }
    \label{fig:alpha-gamma-tuning}
\end{figure*}

Figure~\ref{fig:alpha-gamma-tuning} (a) shows the results of the experiment of the parameter sensitivity with respect to $\alpha$.
WS-CMA-ES shows better performance than the original CMA-ES with any $\alpha$ value in the experiment.
Therefore, WS-CMA-ES is robust against the change in $\alpha$ value.

Next, we conducted experiments of the parameter sensitivity with respect to $\gamma$.
This parameter has the identical trade-off as well.
While extremely large $\gamma$ will lead a promising distribution to uniform distribution, extremely small $\gamma$ will lead a promising distribution to a promising distribution overly fitted to a prior task.
In this experiment, we fixed $\alpha=0.1$ and varied $\gamma$ from $0.05$ to $0.25$ in increments of $0.05$.

Figure~\ref{fig:alpha-gamma-tuning} (b) shows the result of the experiment of the parameter sensitivity with respect to $\gamma$.
WS-CMA-ES shows better performance than the original CMA-ES in any settings of $\gamma$ value and it implies the robustness of WS-CMA-ES with respect to the $\gamma$ value as well.
However, as the $\gamma$ value increases, the performance becomes slightly worse.
We recommend a bit smaller $\gamma$ value ($\approx 0.1$) based on this result.

\subsection{Problem with Task Dissimilarity}
In this section, we investigate the effects of $\alpha$ and $\gamma$ on problems with task dissimilarity.
We apply WS-CMA-ES with various values of $\alpha$ and $\gamma$ to Rotated Ellipsoid Function with offset $b = 0.2$, which is a case with \emph{task dissimilarity} (Section 4.2).

\begin{figure*}[t]
    \centering
    \hspace*{\fill}
    \subfloat[][The parameter sensitivity of WS-CMA-ES with respect to $\alpha$]{
        \includegraphics[width=0.45\linewidth]{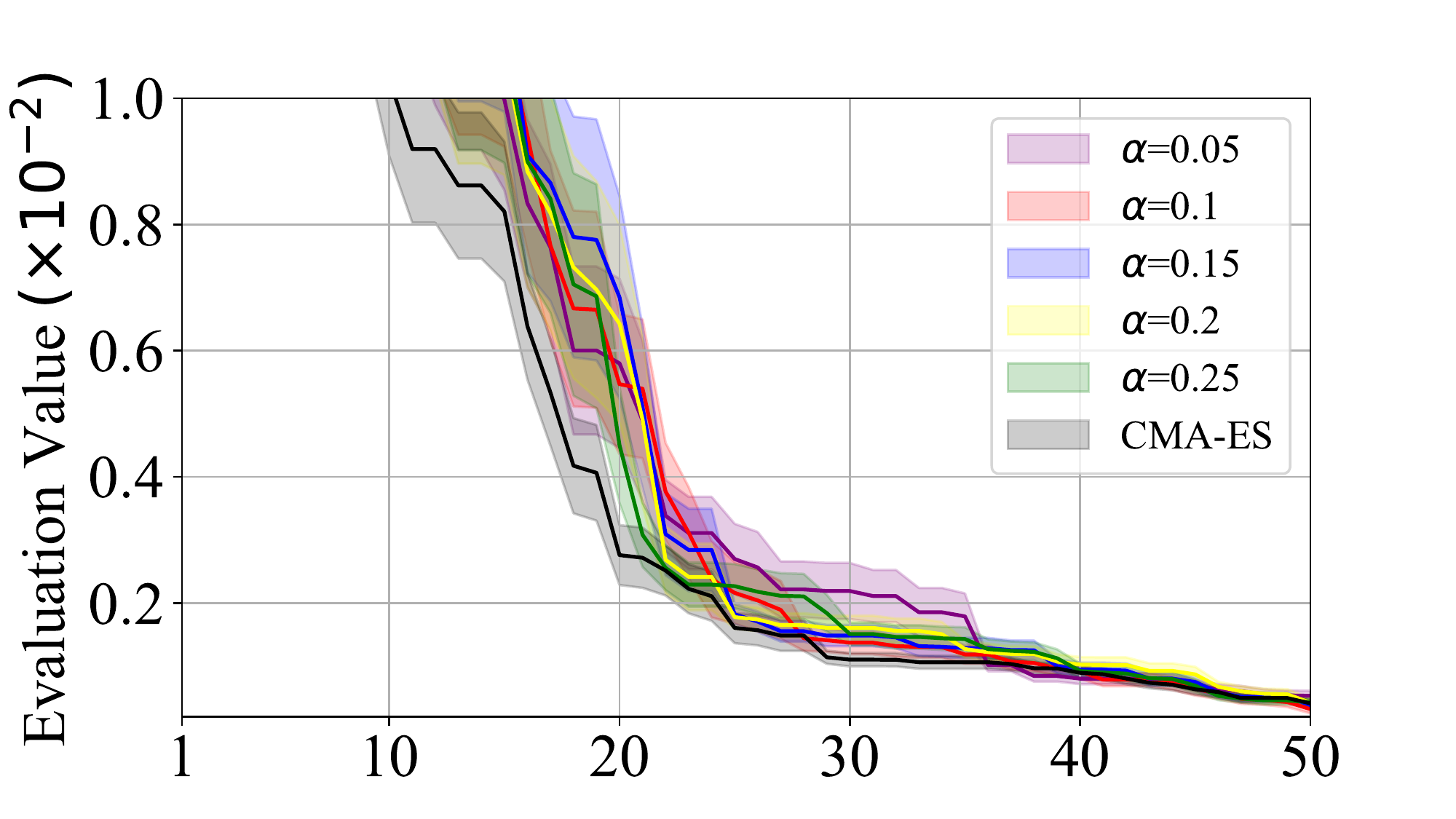}
    }
    \hspace*{\fill}
    \subfloat[][The parameter sensitivity of WS-CMA-ES with respect to $\gamma$ \label{subfig:gamma}]{
        \includegraphics[width=0.45\linewidth]{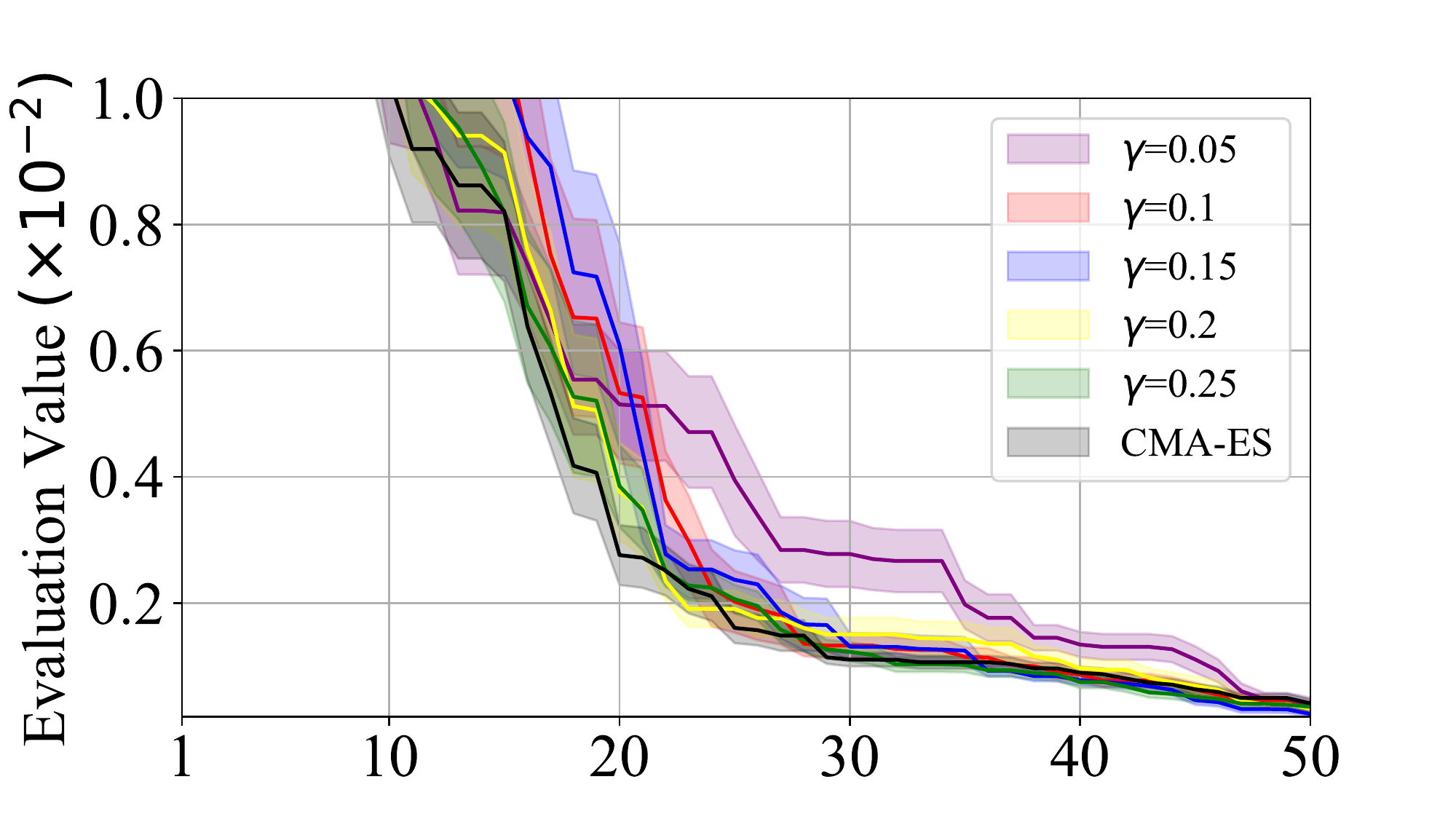}
    }
    \hspace*{\fill}
    \vspace{0mm}
    \caption{Experiments to clarify the parameter sensitivity of the WS-CMA-ES with respect to the control parameters $\alpha$ and $\gamma$ for the task-dissimilar problem.}
    \label{fig:alpha-gamma-task-dissimilarity}
\end{figure*}

Figure~\ref{fig:alpha-gamma-task-dissimilarity} (a) shows the results of the experiment of the parameter sensitivity with respect to $\alpha$.
In this task-dissimilar problem, the CMA-ES shows a bit better performance than the WS-CMA-ES, which is consistency with the result in Section 4.2.
While the performance of the WS-CMA-ES is robust with respect to $\alpha$ overall, the WS-CMA-ES with $\alpha = 0.05$ shows a slightly unstable behavior.
This is because that $\alpha$ has the effect of regularization and WS-CMA-ES with a small $\alpha$ overtrusts the information in the source task.

Figure~\ref{fig:alpha-gamma-task-dissimilarity} (b) shows the results of the experiment of the parameter sensitivity with respect to $\gamma$.
As in case of the experiment for $\alpha$, the CMA-ES shows better performance than the WS-CMA-ES, which is the consistency result.
While the performance of the WS-CMA-ES is robust with respect to $\gamma$ overall, the WS-CMA-ES with $\gamma = 0.05$ shows an unstable behavior.
This is because if the setting of $\gamma$ is too small, the promising region estimates will be overfitted to the source task.

In conclusion, proper settings of $\alpha$ and $\gamma$ work well for the task-dissimilar problem.
Specifically, the parameters $\alpha = 0.1$ and $\gamma = 0.1$ used in the experiments in the main paper showed robust performance in this experiment as well.
Therefore, we recommend these settings even if there is no prior information on the similarity between the source task and the target task.

